%% file: main.tex
\PassOptionsToPackage{numbers,sort&compress}{natbib}
\documentclass{article}
\usepackage{arxiv}
\usepackage{booktabs}
\usepackage{multirow} 
\usepackage{amsmath}
\usepackage{amssymb}
\usepackage{graphicx}
\usepackage{wrapfig}
\usepackage{multirow}
\usepackage{tabularx}
\usepackage{natbib} 

\usepackage{xcolor}

\title{Event-VLA: Action-Conditioned Event Fusion for Robust Vision-Language-Action Model}
\date{}

\newcommand*\samethanks[1][\value{footnote}]{\footnotemark[#1]}

\author{%
 \\
  \textbf{Jiaxin Liu\textsuperscript{1}\thanks{These authors contributed equally to this work.},\ 
  Xun Xu\textsuperscript{1}\samethanks[1],\ 
  Zhenhao Zhang\textsuperscript{1}\samethanks[1],\ 
  Hanqing Wang\textsuperscript{2},\
  Ruiqi Chen\textsuperscript{3}, }\\
  \textbf{Shi Chang\textsuperscript{4},\ 
  Weiyu Guo\textsuperscript{2},\ 
  Laurent Kneip\textsuperscript{1}}\thanks{Corresponding author.}\\[6pt]
  \textsuperscript{1}ShanghaiTech\quad
  \textsuperscript{2}HKUST(GZ)\quad
  \textsuperscript{3}UMich\quad
  \textsuperscript{4}SJTU\\[3pt]
  \texttt{\{liujx2024, xuxun2024, zhangzhh2024, lkneip\}@shanghaitech.edu.cn}\\
  \texttt{hwang201@connect.hkust-gz.edu.cn\quad ruiqich@umich.edu}\\
  \texttt{stellarlane@sjtu.edu.cn\quad guoweiyu96@gmail.com}
}

\begin{document}

\maketitle

\input{sec/0abstract}

\input{sec/1intro}

\input{sec/2related_works}
\input{sec/3methodology}

\input{sec/4experiments}

\input{sec/5conclusion}
\clearpage
\bibliographystyle{unsrtnat}
\bibliography{example}
\clearpage
\input{appendix/appendix}

\end{document}

%% file: sec/0abstract.tex
\begin{abstract}
Vision-Language-Action (VLA) models have become an important paradigm of embodied AI. However, existing VLA models typically assume well-lit and stable indoor settings, while real-world embodied manipulation may involve degraded RGB observations caused by illumination shifts, posing critical challenges for robust robotic manipulation. To address this gap, we propose \textbf{Event-VLA}, an event-enhanced VLA framework for generalizable manipulation across varying illumination conditions. We formulate VLA-based manipulation under degraded visibility as a practical robustness problem for RGB-centric policies, and introduce event streams as an illumination-robust, motion-sensitive complementary observation to improve robustness across visibility levels. Specifically, unlike conventional multimodal fusion that directly merges event features into the global semantic token space, Event-VLA injects event information through an action-query routing pathway. It uses learnable action queries to extract task-relevant semantics from the VLA reasoning process, and selectively aggregates event tokens via gated cross-attention to construct event-aware action representations. This design preserves the pretrained RGB-language semantic priors while effectively leveraging event information for robust action prediction. Experiments in simulation and real-world deployment show that Event-VLA maintains strong manipulation performance under normal lighting and improves success rates under low-light degradation and near-dark real-world settings.

\end{abstract}

\keywords{Vision-Language-Action, Robot Manipulation, Event Camera}

%% file: sec/1intro.tex
\section{Introduction}
\label{sec:introduction}

Vision-language-action (VLA) models~\citep{zhang2026dreamvla,liu2025rdt,wen2025dexvla,zhao2025cot,zheng2025tracevla,tian2025predictive,zhang2025up,lin2025hif,ma2026survey,zhang2026unihm} have emerged as an important foundation for general-purpose robotic manipulation by mapping language instructions and visual observations to robot actions. 
Recent action generation paradigms, represented by flow matching-based policies~\cite{black2024pi_0,intelligence2025pi_,intelligence2025pi}, together with unified multimodal representation learning approaches~\cite{bu2025univla,cen2025worldvla,liang2025mm,deng2025emerging,yang2026mmada,zhao2025unified,team2025gemini,chi2025diffusion,shukor2025smolvla}, have further improved the capability of VLAs in complex task planning and continuous action generation. However, most existing VLAs are built upon idealized visual observations, implicitly assuming well-illuminated manipulation scenes, stable imaging conditions, and high-quality RGB inputs. In real-world robotic manipulation, this premise is often violated by illumination changes, sensor noise, and motion blur, leading to degraded semantic grounding and unstable action prediction~\cite{fei2025libero}. Consequently, robust manipulation in open real-world environments remains challenging.

\begin{figure}[htbp]
    \centering
    \includegraphics[width=1\linewidth]{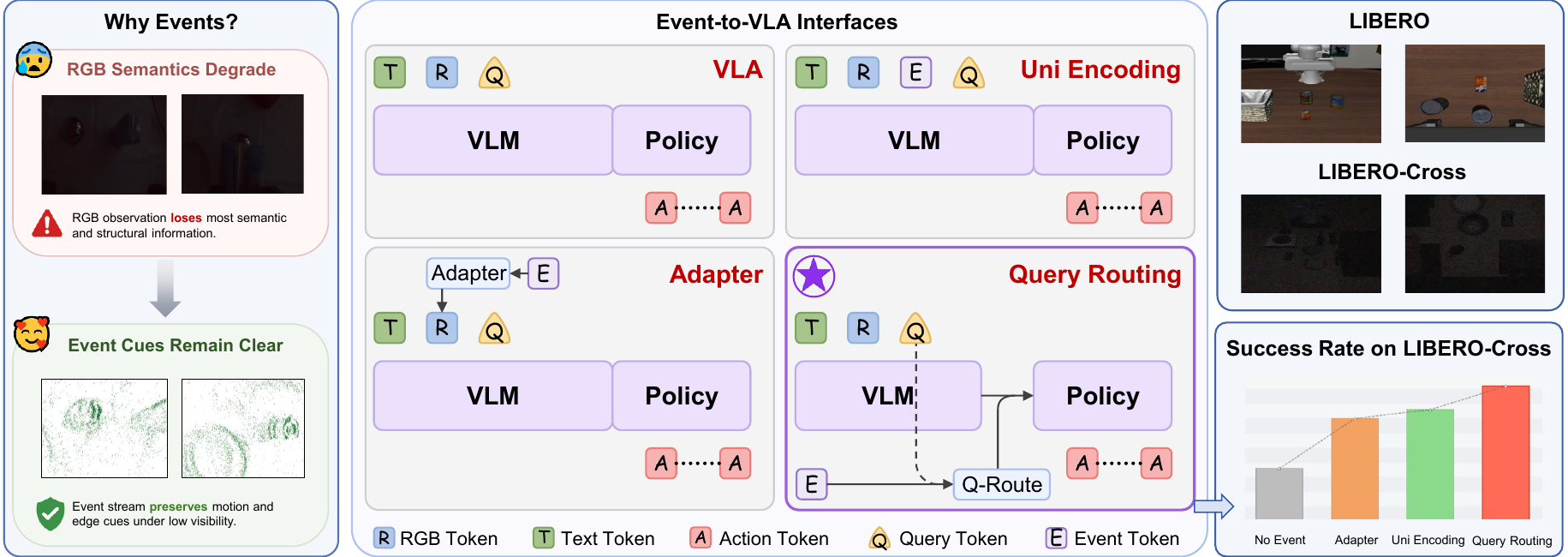}
    \caption{
    Event-to-VLA interface comparison. Under degraded visibility, events retain motion and edge cues when RGB observations degrade. Action-query routing injects event information into the action pathway more effectively than direct token merging or adapter fusion.
    }
    \label{fig:teaser}
\end{figure}

Meanwhile, event cameras provide an event-driven sensing paradigm that asynchronously encodes per-pixel changes in log intensity rather than absolute brightness~\cite{lichtsteiner2008128,brandli2014240}. This temporal-contrast mechanism endows event cameras with high dynamic range, low latency, and strong sensitivity to motion and illumination changes, enabling them to capture stable dynamic visual cues under challenging lighting and fast-motion conditions~\cite{gallego2020event,chakravarthi2024recent}. Event streams therefore serve as a light-robust and motion-sensitive complementary observation for VLAs, compensating for action-time dynamic cues that are often weakened or corrupted in degraded RGB observations~\cite{zheng2023deep}. This raises a key interface question: \textbf{how should asynchronous high-frequency event streams be incorporated into pretrained VLA models?}

To address this question, we propose \textbf{Event-VLA}, an event-enhanced framework for robotic manipulation under degraded visibility. The core design principle of Event-VLA is that \textbf{semantic context should be modeled by the pretrained VLA, while event features should be injected into the action pathway in a controlled manner}. Specifically, Event-VLA uses learnable action queries to extract task-relevant semantic representations from the VLA reasoning process and selectively receives event tokens through gated cross-attention to form event-aware action representations. This design avoids directly merging event tokens into the global semantic token space. As a result, Event-VLA leverages event streams to improve robustness under degraded visibility while preserving the RGB-language semantic priors of pretrained VLAs as much as possible.

We evaluate Event-VLA in both simulation and real-world robotic manipulation. Experimental results show that Event-VLA substantially improves robustness under degraded visibility compared with RGB-only VLAs and multiple event fusion strategies, while maintaining performance under normal visibility. Ablation studies further show that the benefit of events depends not only on the representation, but also on the interface through which events are incorporated into pretrained VLAs.

The main contributions of this work are as follows:
\begin{itemize}

    \item We formulate degraded-visibility VLA manipulation as an event-to-VLA interface problem, and introduce PREI, a lightweight decomposition of event activity into action-time motion residuals.

    \item We propose \textbf{Event-VLA}, an action-conditioned interface that keeps event tokens outside the pretrained semantic backbone and routes them to action representations through gated cross-attention and query-guided routing.

    \item We show through LIBERO, LIBERO-Cross, interface ablations, and real-world Franka tasks that the proposed event-to-action interface improves low-light robustness while preserving normal-light VLA performance.
\end{itemize}

%% file: sec/2related_works.tex
\begin{figure}[htbp]
    \centering
    \includegraphics[width=1\linewidth]{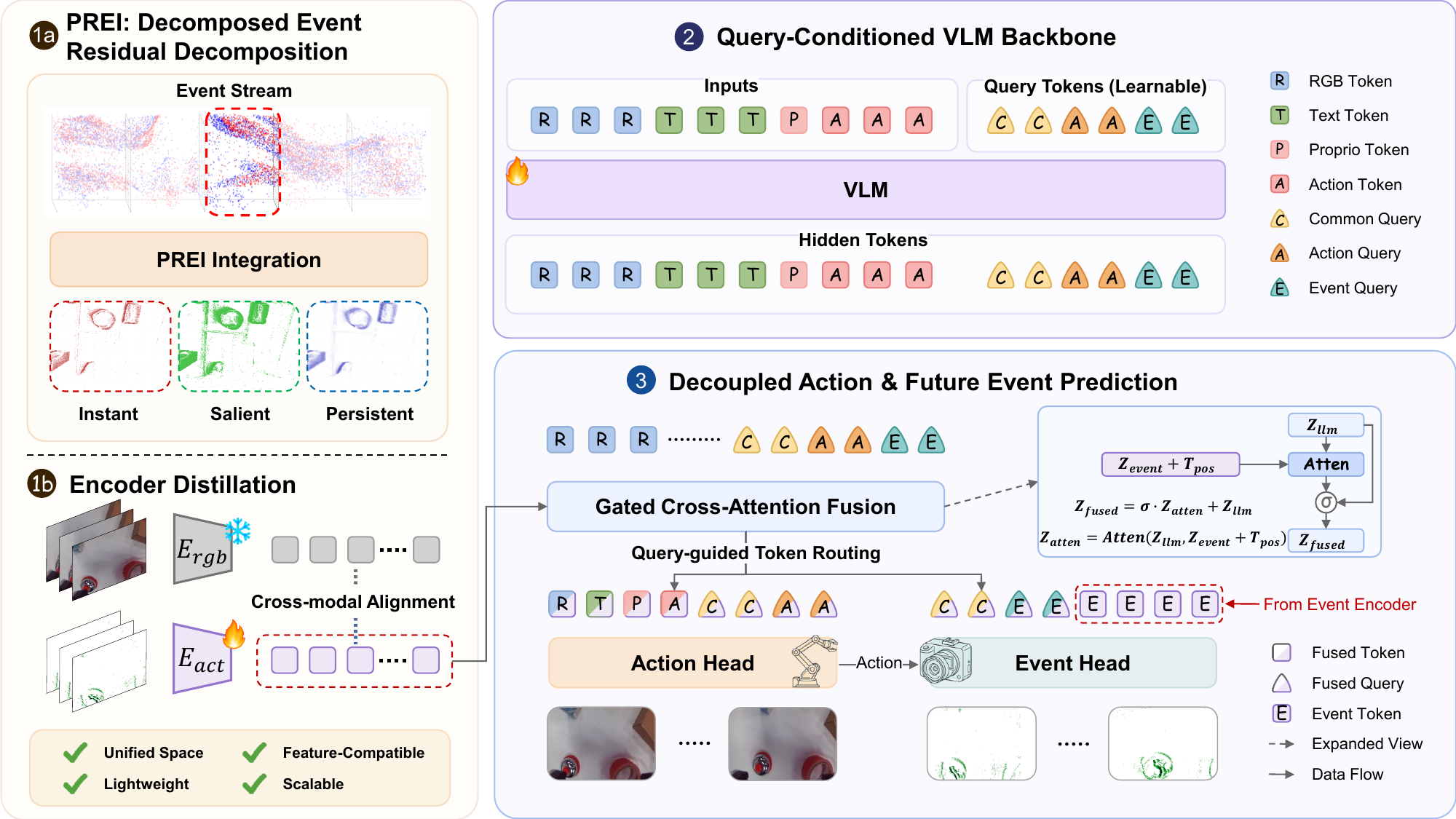}
    \caption{
    Overview of Event-VLA. Event streams are first compressed into PREI residual maps and encoded as event tokens. The pretrained VLA builds the RGB-language-proprioceptive semantic context with query tokens. Event residuals are then selectively injected into action slots through an action-conditioned gated cross-attention interface.
    }
    \label{fig:pipeline}
\end{figure}
\section{Related Work}
\label{sec:related_work}

\textbf{Vision-Language-Action Models}.
Vision-language-action (VLA) models map language instructions and visual observations to executable robot actions and have become a major paradigm for general-purpose robotic manipulation. Autoregressive VLAs~\cite{brohan2022rt,zitkovich2023rt,kim2024openvla,alayrac2022flamingo} formulate action prediction as sequential token generation, while diffusion- and flow-based policies~\cite{team2024octo,black2024pi_0,intelligence2025pi_,intelligence2025pi} generate continuous action trajectories through denoising or flow matching. Beyond these action-generation paradigms, recent VLA models further explore unified multimodal token spaces, shared latent contexts, and stronger action representations~\cite{bu2025univla,cen2025worldvla,zhang2026dreamvla,liang2025mm,liu2025rdt,wen2025dexvla,zhao2025cot,zheng2025tracevla,tian2025predictive,zhang2025up,lin2025hif,bi2026vla}, improving instruction following, task generalization, and long-horizon manipulation~\cite{driess2023palm,deng2025stereovla,huang2025tactile,cheng2025omnivtla}. However, most existing VLAs remain primarily grounded in frame-based RGB observations, whose visual evidence can degrade under low illumination, sensor noise, motion blur, or occlusion.

\textbf{Event-based Robotic Perception and Manipulation}. Event cameras asynchronously capture brightness changes with high temporal resolution and high dynamic range, making them suitable for low-light, high-speed, and motion-blur settings~\cite{gallego2020event,lagorce2016hots}. Event-based perception has been studied for motion estimation~\cite{zhu2018ev}, depth and geometry estimation~\cite{hidalgo2020learning}, and intensity or video reconstruction~\cite{rebecq2019high,rebecq2019events,sironi2018hats,wan2022learning}. In robotics, events have been used for visual servoing, tracking, grasping, and manipulation under challenging sensing conditions~\cite{vinod2025sebvs,hassan2024efficient,li2020event,rebecq2019events,muthusamy2021neuromorphic,guo2024force,sun2022event,liu2025gs}. Recent foundation-model works further connect events with multimodal understanding and embodied policies, including EventGPT~\cite{liu2025eventgpt} and EventVGGT~\cite{ren2026eventvggt}. Event-VLA is complementary to these efforts: it focuses on how event streams should interface with pretrained VLAs, rather than treating events as another globally fused perception modality.

%% file: sec/3methodology.tex
\section{Method}
\label{sec:method}

We instantiate the proposed action-conditioned event interface in \textbf{Event-VLA}. As shown in Fig.~\ref{fig:pipeline}, Event-VLA preserves the pretrained VLA as the semantic backbone and injects event information through action-related pathways. The method consists of three components. First, asynchronous event streams are compressed into physical residual maps and encoded into VLA-compatible event tokens. Second, learnable event queries are fed into the VLM backbone and contextualized by multimodal representations. Third, event tokens are selectively routed to action slots through an action-conditioned interface.

\subsection{Problem Formulation}
\label{sec:problem_formulation}

Given a language instruction $\ell$, an RGB observation $I_t$, and a proprioceptive state $s_t$, a pretrained VLA predicts an action chunk
\begin{equation}
    \hat{A}_t
    =
    \pi_{\theta}(I_t,s_t,\ell),
    \qquad
    \hat{A}_t=
    [\hat{a}_t,\ldots,\hat{a}_{t+K-1}]
    \in\mathbb{R}^{K\times d_a},
\label{eq:rgb_vla_policy}
\end{equation}
where $K$ is the action horizon and $d_a$ is the action dimension. Under degraded visibility, RGB observations may provide unreliable texture, color, and boundary cues, while manipulation still depends on local physical changes around the robot, objects, and contact regions.

We augment the pretrained VLA with an event history $\mathcal{S}_{t-H:t}$. Each event is represented as
$e_i=(x_i,y_i,\tau_i,p_i)$, where $(x_i,y_i)$ is the pixel location, $\tau_i$ is the timestamp, and $p_i$ is the polarity. The event-enhanced policy is written as
\begin{equation}
    \hat{A}_t
    =
    \pi_{\theta,\phi}
    (I_t,s_t,\ell,\mathcal{S}_{t-H:t}),
\label{eq:event_policy}
\end{equation}
where $\phi$ denotes event-side parameters. Event-VLA aims to use the action-time physical residuals in $\mathcal{S}_{t-H:t}$ while preserving the global semantic context of the pretrained VLA.

\subsection{PREI: Decomposing Events into Action-Time Physical Residuals}
\label{sec:event_tokens}

Raw event streams are sparse, asynchronous, and irregularly sampled. We use
\textbf{PREI} (\emph{Physical Residual Event Integration}) as a lightweight
representation layer that converts the event history $\mathcal{S}_{t-H:t}$ into
compact physical residual maps. For a pixel location $u=(x,y)$ and decay
constant $\tau$, PREI first computes a polarity-agnostic activity map~\cite{alonso2019ev}:
\begin{equation}
    A_{\tau}(u)
    =
    \sum_{i:(x_i,y_i)=u}
    \exp\left(-\frac{t-\tau_i}{\tau}\right),
\label{eq:prei_activity}
\end{equation}
where $t$ is the current prediction time, $\tau_i$ is the timestamp of event
$e_i$, and the summation is taken over events at pixel $u$ within the event
window. Thus PREI forms a three-channel residual map:
\begin{equation}
\begin{gathered}
E_t^{\mathrm{prei}}
=
[E_t^{\mathrm{ins}}, E_t^{\mathrm{sal}}, E_t^{\mathrm{per}}]
\in [0,1]^{H_{\mathrm{img}}\times W_{\mathrm{img}}\times 3},\\[2pt]
E_t^{\mathrm{ins}}(u)
=
\rho\!\left(\frac{A_{\tau_{\mathrm{ins}}}(u)}{\kappa_{\mathrm{ins}}}\right),
\quad
E_t^{\mathrm{sal}}(u)
=
\rho\!\left(\frac{A_{\tau_{\mathrm{sal}}}(u)}
{\eta B_{\tau_{\mathrm{sal}}}(u)+\epsilon}\right),
\quad
E_t^{\mathrm{per}}(u)
=
\rho\!\left(\frac{C(u)}{\kappa_{\mathrm{per}}}\right).
\end{gathered}
\label{eq:prei_channels}
\end{equation}
Here, $B_{\tau}(u)=(G_{\sigma}*A_{\tau})(u)$ denotes the locally smoothed
activity map, where $G_{\sigma}$ is a smoothing kernel and $*$ denotes
convolution. $\tau_{\mathrm{ins}}$ and $\tau_{\mathrm{sal}}$ are decay constants,
$\kappa_{\mathrm{ins}}$ and $\kappa_{\mathrm{per}}$ are scale factors,
$C(u)=\sum_{i:(x_i,y_i)=u}1$ counts events at pixel $u$ within the event window,
and $\eta$ and $\epsilon$ control local normalization and numerical stability.
PREI decomposes event activity into three action-time residual cues: instantaneous changes for recent robot-object motion, salient changes for locally prominent activity around manipulated objects, and persistent traces for short-term contours when RGB texture and color degrade. Normalization constants and implementation details are provided in Appendix~\ref{app:prei_details}.

An event encoder maps each PREI map into VLA-compatible event tokens:
\begin{equation}
    Y_t^e = \Phi_e(E_t^{\mathrm{prei}}) = [g_t^e; P_t^e],
\label{eq:event_tokens}
\end{equation}
where $g_t^e$ denotes the global event token and $P_t^e$ denotes
spatial event tokens. We use feature-level distillation from a frozen visual
teacher to improve compatibility with the VLA hidden space, while keeping event
tokens outside the global semantic context of the pretrained VLA. For feature distillation, we use native RGB-event pairs from N-ImageNet~\cite{kim2021n} and generate RGB-event pairs from LIBERO~\cite{liu2023libero} videos using v2e~\cite{hu2021v2e}. More detail could be found in \ref{app:event_encoder}

\subsection{Action-Conditioned Event Interface}
\label{sec:action_conditioned_interface}

Event-VLA introduces event information only after the pretrained VLA has formed its RGB-language-proprioceptive-action context. The VLA input contains RGB visual tokens, a proprioceptive token, language tokens, action placeholders, and learnable query tokens:
\begin{equation}
    X_t=
    [Z_t^v;\;z_t^s;\;Z^{\ell};\;Z_0^a;\;Q_0^c;\;Q_0^a;\;Q_0^e],
\label{eq:vla_input}
\end{equation}
where $Q_0^c,Q_0^a,Q_0^e$ denote common, action, and event queries, respectively. The pretrained VLA backbone produces contextualized hidden states:
\begin{equation}
    H_t=F_{\mathrm{vla}}(X_t),
    \qquad
    H_t=[Z_t^{\mathrm{lm}};\;Q_t^c;\;Q_t^a;\;Q_t^e],
\label{eq:vla_forward}
\end{equation}
where $Z_t^{\mathrm{lm}}$ denotes the contextualized tokens, and
$Q_t^c,Q_t^a,Q_t^e$ are the contextualized query states.

The event encoder produces event tokens $Y_t^e$, which are not passed into
$F_{\mathrm{vla}}$. Instead, they are fused with the VLA hidden tokens through gated cross-attention:
\begin{equation}
    Z_t^{\mathrm{attn}}
    =
    \operatorname{CrossAttn}
    \left(
        W_q Z_t^{\mathrm{lm}},
        W_k(Y_t^e+T_{\mathrm{pos}}),
        W_v(Y_t^e+T_{\mathrm{pos}})
    \right),
\label{eq:event_cross_attn}
\end{equation}
\begin{equation}
    \Gamma_t
    =
    \sigma
    \left(
        f_{\mathrm{gate}}
        (Z_t^{\mathrm{lm}},Q_t^c,Q_t^a,Q_t^e,Z_t^{\mathrm{attn}})
    \right),
    \qquad
    Z_t^{\mathrm{fused}}
    =
    \operatorname{Norm}
    \left(
        Z_t^{\mathrm{lm}}
        +
        \Gamma_t\odot Z_t^{\mathrm{attn}}
    \right).
\label{eq:gated_event_injection}
\end{equation}

The fused tokens are then separated by query-guided token routing. The common and action queries route action-relevant tokens to the action head, while the common and event queries route event-relevant tokens to the auxiliary event head:
\begin{equation}
    Z_t^{\mathrm{act}}
    =
    \operatorname{Route}_{\mathrm{act}}
    (Z_t^{\mathrm{fused}},Q_t^c,Q_t^a),
    \qquad
    Z_t^{\mathrm{evt}}
    =
    \operatorname{Route}_{\mathrm{evt}}
    (Z_t^{\mathrm{fused}},Q_t^c,Q_t^e).
\label{eq:query_guided_routing}
\end{equation}
The action head predicts the future action chunk, and the event head predicts future PREI event targets conditioned on the routed event tokens and the action representation:
\begin{equation}
    \hat{A}_t
    =
    f_{\mathrm{act}}(Z_t^{\mathrm{act}}),
    \qquad
    \hat{E}^{\mathrm{prei}}_{t+1:t+K}
    =
    f_{\mathrm{evt}}(Z_t^{\mathrm{evt}},\hat{A}_t).
\label{eq:action_event_decode}
\end{equation}

Since $Y_t^e$ is never passed into $F_{\mathrm{vla}}$, event information does not participate in the global semantic self-attention of the pretrained VLA. Instead, event tokens affect downstream action and future-event prediction only through post-backbone gated fusion and query-guided routing. This structural separation distinguishes Event-VLA from shared-context event fusion methods that directly concatenate event tokens with RGB-language tokens.
\subsection{Training Objective and Inference}
\label{sec:training_inference}

The primary training objective is action prediction. Given a ground-truth action chunk
$A_t=\{a_t,\ldots,a_{t+K-1}\}$ and the predicted action chunk
$\hat{A}_t=\{\hat{a}_t,\ldots,\hat{a}_{t+K-1}\}$, we use an $\ell_1$ action loss:
\begin{equation}
    \mathcal{L}_{\mathrm{act}}
    =
    \frac{1}{K}
    \sum_{k=0}^{K-1}
    \left\|
        \hat{a}_{t+k}
        -
        a_{t+k}
    \right\|_1 .
\label{eq:action_loss}
\end{equation}

The auxiliary event head is supervised by future PREI event targets. Let
$E^{\mathrm{prei}}_{t+k}$ denote the ground-truth future PREI target and
$\hat{E}^{\mathrm{prei}}_{t+k}$ denote the event-head prediction. We use a content mask
$M_{t+k}\in[0,1]^{H\times W}$ to compute the event loss only on valid content regions:
\begin{equation}
    \mathcal{L}_{\mathrm{evt}}
    =
    \frac{1}{K}
    \sum_{k=1}^{K}
    \frac{
        \left\|
            M_{t+k}
            \odot
            \left(
                \hat{E}^{\mathrm{prei}}_{t+k}
                -
                E^{\mathrm{prei}}_{t+k}
            \right)
        \right\|_1
    }{
        \left\|M_{t+k}\right\|_1+\epsilon
    } .
\label{eq:event_loss}
\end{equation}
Here $M_{t+k}$ is broadcast to all PREI channels, and $\epsilon$ is a small constant for numerical stability.

To further regularize the event prediction, we also impose a first-order derivative consistency loss between the predicted future event map and the ground-truth future event map:
\begin{equation}
    \mathcal{L}_{\mathrm{deriv}}
    =
    \frac{1}{K}
    \sum_{k=1}^{K}
    \frac{
        \left\|
            M_{t+k}
            \odot
            \left(
                \nabla \hat{E}^{\mathrm{prei}}_{t+k}
                -
                \nabla E^{\mathrm{prei}}_{t+k}
            \right)
        \right\|_1
    }{
        \left\|M_{t+k}\right\|_1+\epsilon
    } ,
\label{eq:event_derivative_loss}
\end{equation}
where $\nabla$ denotes the first-order finite-difference derivative operator on the PREI representation.

The full training objective is
\begin{equation}
    \mathcal{L}
    =
    \mathcal{L}_{\mathrm{act}}
    +
    \lambda_{\mathrm{evt}}
    \mathcal{L}_{\mathrm{evt}}
    +
    \lambda_{\mathrm{deriv}}
    \mathcal{L}_{\mathrm{deriv}} .
\label{eq:full_loss}
\end{equation}
The two event-head losses, $\mathcal{L}_{\mathrm{evt}}$ and
$\mathcal{L}_{\mathrm{deriv}}$, are used as regularization terms. They encourage the fused and event-routed representations to preserve transient physical information from the event stream, but they are optional during inference.

%% file: sec/4experiments.tex
\section{Experiments}
\label{sec:experiments}

Our experiments center around four primary questions:
\textbf{Q1:} Does Event-VLA maintain pretrained VLA performance under normal visibility and improve robustness under progressive visibility degradation?
\textbf{Q2:} Does the proposed event interface preserve the original RGB-language action pathway of the pretrained VLA when event inputs are absent?
\textbf{Q3:} What is the right representation and interface for event streams in pretrained VLAs?
\textbf{Q4:} How do PREI, query-conditioned routing, and auxiliary event regularization contribute to the final policy performance?
We evaluate these questions on the original LIBERO benchmark, a controlled low-visibility extension named LIBERO-Cross, and a real-world Franka deployment.

\subsection{Progressive Visibility Degradation Benchmark}
\label{sec:degradation_benchmark}

We evaluate Event-VLA on \textbf{LIBERO}~\cite{liu2023libero} and its low-visibility extension, \textbf{LIBERO-Cross}. LIBERO-Cross preserves the original tasks, language instructions, robot states, and evaluation protocol, while applying controlled RGB degradations at three levels: LL-Mild, LL-Dark, and LL-Severe. These levels vary luminance attenuation, signal-to-noise ratio at a reference $18\%$ gray level, and weak motion blur, enabling controlled evaluation of VLA robustness as frame-based observations become unreliable.

We generate synchronized event-side inputs with an efficient RGB-to-event simulator trained from pseudo-event supervision on LIBERO videos~\cite{hu2021v2e,gehrig2021dsec,rebecq2018esim,li2024blinkvision,zhang2024v2ce,rebecq2019high}. The simulator predicts compact event representations used by Event-VLA. Details of the degradation model, simulator, and validation are provided in Appendix~\ref{app:libero_cross}.

\subsection{Training and Evaluation Setup}
\label{sec:exp_setup}

We adopt Prismatic-7B ~\cite{karamcheti2024prismatic} as the backbone, and initialize it with weights from OpenVLA, which were pretrained on OXE ~\cite{o2024open}. All other modules are randomly initialized. We train separate models for each LIBERO sub-benchmark with batch size 64, action chunk size 8. Each model is trained for approximately 140k steps on eight H100 GPUs. More details are provided in Appendix~\ref{app:implementation}.

We compare Event-VLA with representative RGB-based VLA policies, including OpenVLA, OpenVLA-OFT, $\pi_0$, and MM-ACT, etc. For low-visibility evaluation, we include event-interface baselines: unified event token encoding, RGB-Event adapter fusion, and query routing fusion. We report success rate (SR) for each  sub-benchmark and the average SR across sub-benchmarks.

\subsection{Main Results under Progressive Visibility Degradation}
\label{sec:main_results}

\begin{table}[htbp]
\centering
\caption{Success rate (\%) on the original LIBERO benchmark under normal visibility. ``Ours w/o event'' denotes the trained Event-VLA with its event pathway disabled at inference.}
\label{tab:libero_normal}
\setlength{\tabcolsep}{6pt}
\begin{tabular*}{\linewidth}{@{\extracolsep{\fill}}lccccc}
\toprule
Method & Spatial & Object & Goal & Long & Avg. \\
\midrule
OpenVLA  & 84.7 &  88.4 &  79.2 & 53.7 & 76.5 \\
$\pi_0$  & \underline{96.8} & \underline{98.8} &  95.8 & 85.2 & 94.2 \\
OpenVLA-OFT&   96.2 & 98.3 & \underline{96.2} & 90.7 & 95.4 \\
ResVLA &  96.8 &  98.6 &  \textbf{97.4} & 92.4 &  \underline{96.3} \\
MM-ACT & \textbf{97.8} & \textbf{99.4} & 94.8 & \underline{93.0} & \underline{96.3} \\
\midrule
Ours w/o event&  94.4 & 99.2 & 96.8 & 94.4 & 96.2 \\
Ours & 94.2 & \textbf{99.4} & \textbf{97.4} & \textbf{94.8} & \textbf{96.5 }\\
\bottomrule
\end{tabular*}
\end{table}

\begin{table*}[htbp]
\centering
\caption{Success rate (\%) on LIBERO-Cross under progressive visibility degradation.}
\label{tab:libero_dark}
\setlength{\tabcolsep}{6pt}
\begin{tabular*}{\linewidth}{@{\extracolsep{\fill}}lcccccc}
\toprule
Level & Method & Spatial & Object & Goal & Long & Avg. \\
\midrule
\multirow{5}{*}{LL-Mild}
& $\pi_0$ & 94.0 & 96.2 &  93.6 & 82.8 &  91.7 \\
& OpenVLA-OFT & 93.2 & 97.4 & \underline{96.4} & 88.6 & 93.9 \\
& MM-ACT & \textbf{95.8} & \underline{99.2} & 95.6 & \textbf{92.8} & \underline{95.9} \\
& Ours & \underline{94.4} & \textbf{99.4} & \textbf{99.0} & \underline{91.6} & \textbf{96.1} \\
\midrule
\multirow{5}{*}{LL-Dark}
& $\pi_0$ & 89.6 & 94.4 & 91.4 & 81.6 & 89.3 \\
& OpenVLA-OFT & 91.6 & 95.8 & 93.2 & \underline{85.8} & 91.6 \\
& MM-ACT & \textbf{94.4} & \underline{97.6} & \underline{96.8} & 83.2 & \underline{93.0} \\
& Ours & \underline{94.2} & \textbf{98.8} & \textbf{98.6} & \textbf{94.6} & \textbf{96.5} \\
\midrule
\multirow{5}{*}{LL-Severe}
& $\pi_0$ & \underline{62.6} & 76.2  & 52.6 & 56.4 & 61.9 \\
& OpenVLA-OFT & 51.4 & 74.6 & 60.4 & 58.2 & 61.2 \\
& MM-ACT & 58.2 & \underline{76.8} & \underline{65.4} & \underline{78.2} & \underline{69.6} \\
& Ours & \textbf{95.6} & \textbf{97.2} & \textbf{97.8} & \textbf{92.0} & \textbf{95.6} \\
\bottomrule
\end{tabular*}
\end{table*}

\textbf{Experimental setups.} For both LIBERO and LIBERO-Cross, we evaluate each task in every sub-benchmark over 50 rollouts, resulting in 500 evaluation trials per sub-benchmark. We report the success rate (SR) computed over these 500 trials.

We first evaluate whether Event-VLA preserves normal-visibility performance. As shown in Table~\ref{tab:libero_normal}, Event-VLA achieves comparable success rates to RGB-based VLA policies on the original LIBERO benchmark. We further test \emph{Ours w/o event}, where the trained Event-VLA is evaluated with its event pathway disabled. Its comparable performance indicates that the learned event interface does not disrupt the pretrained RGB-language action pathway.

We then evaluate robustness on LIBERO-Cross under LL-Mild, LL-Dark, and LL-Severe degradations. As shown in Table~\ref{tab:libero_dark}, RGB-based policies degrade as visual observations become unreliable, whereas Event-VLA maintains stronger action generation by leveraging event residuals.

\textbf{Results and Analysis.}
Event-VLA improves robustness without disrupting the pretrained VLA pathway. On the original LIBERO benchmark, \emph{Ours w/o event} remains close to \emph{Ours}, indicating that the event interface preserves the RGB-language action pathway. Under progressive degradation, RGB-based baselines drop sharply, especially at LL-Severe, whereas Event-VLA maintains 95.6\% average SR. This suggests that action-conditioned event routing provides  physical cues when RGB observations become unreliable. These results provide empirical evidence for \textbf{Q1} and \textbf{Q2}.

\subsection{Event-to-VLA Interface Ablation}
\label{sec:interface_ablation}

\begin{wraptable}[12]{r}{0.50\linewidth}
\vspace{-4em}
\centering
\caption{Ablation on LIBERO-Cross LL-Severe. $\Delta$ Lat. is latency overhead over no-event (162.208 ms).}
\label{tab:interface_ablation}
% \vspace{-0.4em}
\scriptsize
\setlength{\tabcolsep}{2.5pt}
\renewcommand{\arraystretch}{0.95}

\begin{tabularx}{\linewidth}{lXcc}
\toprule
Type & Variant & SR (\%) & $\Delta$ Lat. (ms)\\
\midrule
\multirow{3}{*}{Repr.}
    & No event & 60.6 & \textbf{+0} \\
    & Time surface & 91.2 & \underline{+2.157} \\
    & \textbf{PREI} & \textbf{95.6} & \underline{+2.157} \\
\midrule
\multirow{3}{*}{Interface}
    & Unified enc. & 95.1 & +62.874 \\
    & RGB/event adapter & 94.2 & \textbf{+1.909} \\
    & \textbf{Query routing} & \textbf{95.6} & \underline{+2.157} \\
\midrule
\multirow{3}{*}{Query}
    & w/o common & 94.5 & +3.708 \\
    & w/o event & 95.2 & \underline{+1.096} \\
    & \textbf{Full queries} & \textbf{95.6} & \textbf{+2.157} \\
\midrule
\multirow{3}{*}{Reg.}
    & None & 94.8 & -- \\
    & w/o mask & 95.1 & -- \\
    & \textbf{Full objective} & \textbf{95.6} & -- \\
\bottomrule
\end{tabularx}
% \vspace{-0.8em}
\end{wraptable}

\textbf{Experimental setups.} We conduct ablations on LIBERO-Cross LL-Severe to study how event cues should interface with pretrained VLA. As shown in Table~\ref{tab:interface_ablation}, we analyze four factors: event representation, fusion interface, query design, and event regularization, and report both success rate (SR) and inference latency overhead ($\Delta$ Lat) on RTX 4090.

\textbf{Results and Analysis.} The ablations reveal that both event representation and interface design are critical under severe degradation. 
Without event input, the average SR drops to 60.6\%, while time surfaces improve it to 91.2\% and PREI further reaches 95.6\%. 
For event-to-VLA fusion, unified event encoding achieves a competitive SR of 95.1\% but incurs much higher latency overhead, whereas query routing obtains the best SR of 95.6\% with only small additional latency. 
The query and regularization ablations further show that full query routing and future-event supervision provide consistent gains. 
Together, these results answer \textbf{Q3} and \textbf{Q4}, showing that Event-VLA's robustness comes from representing events as PREI residuals and routing them through an efficient action-conditioned interface. 
More details are provided in Appendix~\ref{app:ablation_degradation}.

\subsection{Real-World Deployment}
\label{sec:real_world}
\begin{figure}
    \centering
    \includegraphics[width=1\linewidth]{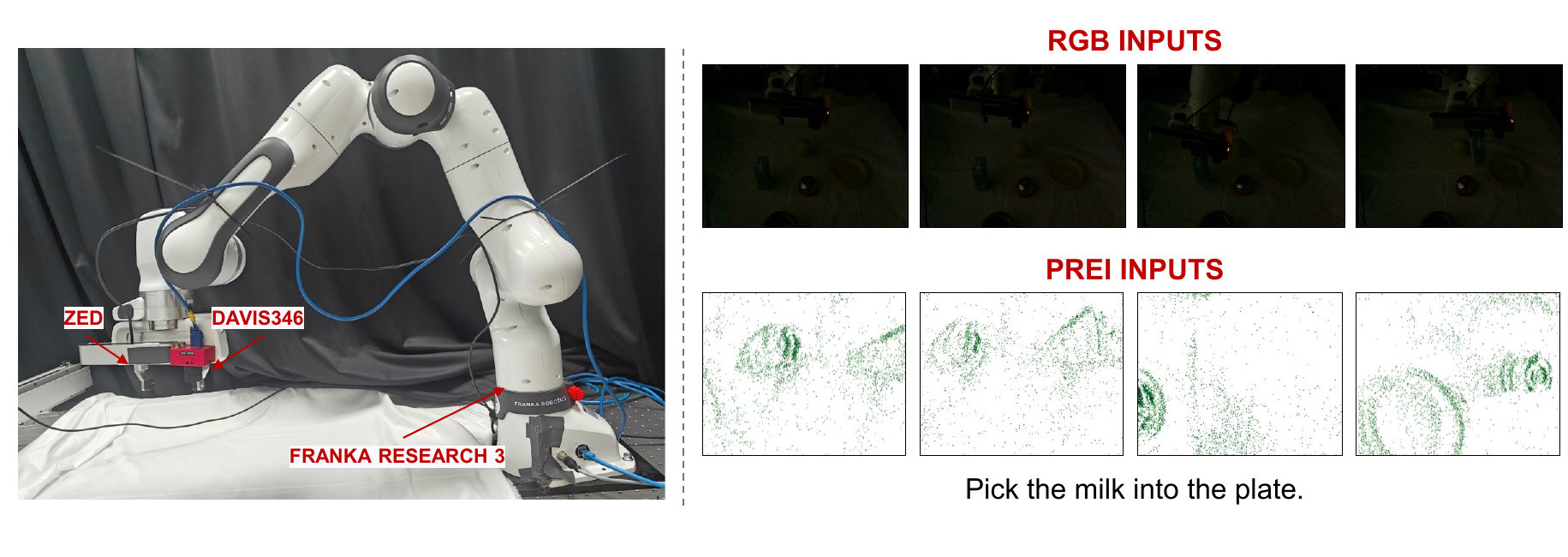}
    \caption{
    Real-world deployment setup and a task under visually degraded conditions with both RGB and Event observation.
    }
    \label{fig:exp}
\end{figure}

\begin{table*}[t]
\centering
\caption{Success rate (\%) on real-world Franka deployment.}
\label{tab:real_world}
\setlength{\tabcolsep}{12pt}
\renewcommand{\arraystretch}{1.05}
\begin{tabular*}{\textwidth}{@{\extracolsep{\fill}}lcccc}
\toprule
Method & Normal & Low-Light & Near-Dark & Avg. \\
\midrule
$\pi_0$ & \textbf{75.0 }& 55.0 & \underline{15.0} & \underline{48.3} \\
OpenVLA-OFT & 70.0 & \underline{57.5} & 12.5 & 46.7 \\
\midrule
Ours w/o queries  & 70 & 65  & 45 & 60 \\
Ours & \underline{72.5} & \textbf{70.0} &\textbf{ 52.5} & \textbf{65.0} \\
\bottomrule
\end{tabular*}
\end{table*}

\textbf{Experimental setups.} We evaluate Event-VLA on a real Franka Research 3 robot with a wrist-mounted ZED camera, an external Orbbec camera, and a DAVIS event camera. We collect 80 demonstrations for four manipulation tasks and evaluate under normal, low-light, and near-dark conditions. The model is deployed on a RTX 4090 server. For real-robot evaluation, we conduct 10 trials for each task under each
lighting condition and report the cumulative success rate across tasks. More details are provided in Appendix~\ref{app:real_robot}.

\textbf{Results and Analysis.} The real-world deployment provides an initial check that the proposed interface remains useful with a physical DAVIS event camera. 
Under normal lighting, it preserves competitive manipulation performance. 
Under visual degradation conditions, Event-VLA is more stable than RGB-based policies.

%% file: sec/5conclusion.tex
\section{Limitations and Conclusion}
\label{sec:conclusion}
\textbf{Conclusion.} We presented \textbf{Event-VLA}, an event-enhanced VLA framework for robust manipulation under degraded visibility. Event-VLA treats event streams as action-time physical residuals rather than global semantic tokens, and injects them into action representations through gated cross-attention and query-guided routing. Experiments on simulation and real-world deployment show that Event-VLA preserves normal-visibility performance while substantially improving robustness under near-dark conditions. Ablations further show that both PREI event representation and the action-conditioned event interface are important for achieving strong robustness with low latency.

% \section{Limitations}
\label{sec:limitation}
\textbf{Limitations.} This work has several limitations. First, LIBERO-Cross relies on an RGB-to-event simulator, which may not fully capture the noise, triggering dynamics, and calibration errors of real event cameras. Second, the real-world evaluation is limited to a small set of tasks and lighting conditions, and broader validation across more objects, environments, and long-horizon tasks is needed. Finally, Event-VLA mainly targets illumination-related visual degradation and requires additional event-camera hardware and synchronization, which may increase deployment complexity.

%% file: appendix/appendix.tex
\appendix

\appendix
\clearpage

\begin{center}
{\Large\bfseries Event-VLA: Action-Conditioned Event Fusion for Robust Vision-Language-Action Model\\ Appendix}
\end{center}
\vspace{1em}

\section{Brief Description of AI Usage}
\label{app:ai_usage}

We used ChatGPT for writing assistance, including grammar checking, style polishing, and phrasing refinement of the manuscript. We also used ChatGPT to obtain suggestions for figure color palettes and visual styling. No figures, experimental results, or scientific claims were generated by AI tools.

\section{Additional Method Details}
\label{app:method_details}

\subsection{PREI Construction Details}
\label{app:prei_details}

PREI converts asynchronous event streams into dense multi-timescale physical residual maps. Given an event window $\mathcal{S}_{t-H:t}$ ending at time $t$, each event is represented as $e_i=(x_i,y_i,\tau_i,p_i)$, where $(x_i,y_i)$ is the pixel location, $\tau_i$ is the timestamp, and $p_i\in\{-1,+1\}$ is the polarity.

Although polarity contains brightness-change direction, our default PREI construction uses polarity-agnostic activity maps. This choice emphasizes the location and temporal intensity of action-induced changes and avoids cancellation between positive and negative events near fast motion boundaries. For a pixel location $u=(x,y)$ and decay constant $\tau$, we define
\begin{equation}
    A_{\tau}(u)
    =
    \sum_{i:(x_i,y_i)=u}
    \exp
    \left(
    -\frac{t-\tau_i}{\tau}
    \right).
\end{equation}

PREI converts the event history into a three-channel residual map:
\begin{equation}
    E_t^{\mathrm{prei}}
    =
    \Psi_{\mathrm{prei}}(\mathcal{S}_{t-H:t})
    =
    [E_t^{\mathrm{ins}},E_t^{\mathrm{sal}},E_t^{\mathrm{per}}]
    \in
    [0,1]^{H_{\mathrm{img}}\times W_{\mathrm{img}}\times 3}.
\end{equation}
The three channels capture complementary temporal aspects of physical residuals.

\paragraph{Instantaneous channel.}
The instantaneous channel emphasizes recent local changes:
\begin{equation}
    E_t^{\mathrm{ins}}(u)
    =
    \rho
    \left(
    \frac{A_{\tau_{\mathrm{ins}}}(u)}
    {\kappa_{\mathrm{ins}}}
    \right),
\end{equation}
where $\tau_{\mathrm{ins}}$ is a short decay constant, $\kappa_{\mathrm{ins}}$ is a scale factor, and $\rho(\cdot)$ is a bounded normalization function.

\paragraph{Salient channel.}
The salient channel highlights locally prominent event activity over a longer time scale:
\begin{align}
    B_{\tau_{\mathrm{sal}}}(u)
    &=
    (G_{\sigma} * A_{\tau_{\mathrm{sal}}})(u), \\
    E_t^{\mathrm{sal}}(u)
    &=
    \rho
    \left(
    \frac{A_{\tau_{\mathrm{sal}}}(u)}
    {\eta B_{\tau_{\mathrm{sal}}}(u)+\epsilon}
    \right),
\end{align}
where $G_{\sigma}$ is a local smoothing kernel and $\epsilon$ avoids division by zero.

\paragraph{Persistent channel.}
The persistent channel preserves event structures accumulated over the full window:
\begin{align}
    C(u)
    &=
    \sum_{i:(x_i,y_i)=u}1, \\
    E_t^{\mathrm{per}}(u)
    &=
    \rho
    \left(
    \frac{C(u)}
    {\kappa_{\mathrm{per}}}
    \right).
\end{align}

Compared with event count images and time surfaces, PREI preserves complementary instantaneous, salient, and persistent activity. Event count images retain accumulated activity but discard temporal recency; time surfaces preserve recent events but may ignore persistent structures. PREI provides a compact residual representation tailored to action-time changes. The factors
$\kappa_{\mathrm{ins}}$ and $\kappa_{\mathrm{per}}$ normalize the instantaneous
and persistent responses, $\eta$ scales the local background, $\epsilon$ avoids
division by zero, and $\rho(\cdot)$ maps each channel to $[0,1]$.

In all experiments, we set the event window to $H=40$ ms, use $\tau_{\mathrm{ins}}=3$ ms and $\tau_{\mathrm{sal}}=10$ ms with a $5\times5$ Gaussian smoothing kernel, set $\eta=0.5$ and $\epsilon=10^{-5}$, use $\rho(\cdot)=\tanh(\cdot)$, and normalize the persistent channel by the 98th percentile of non-zero event counts lower-bounded by 1.

\begin{figure}
    \centering
    \includegraphics[width=1\linewidth]{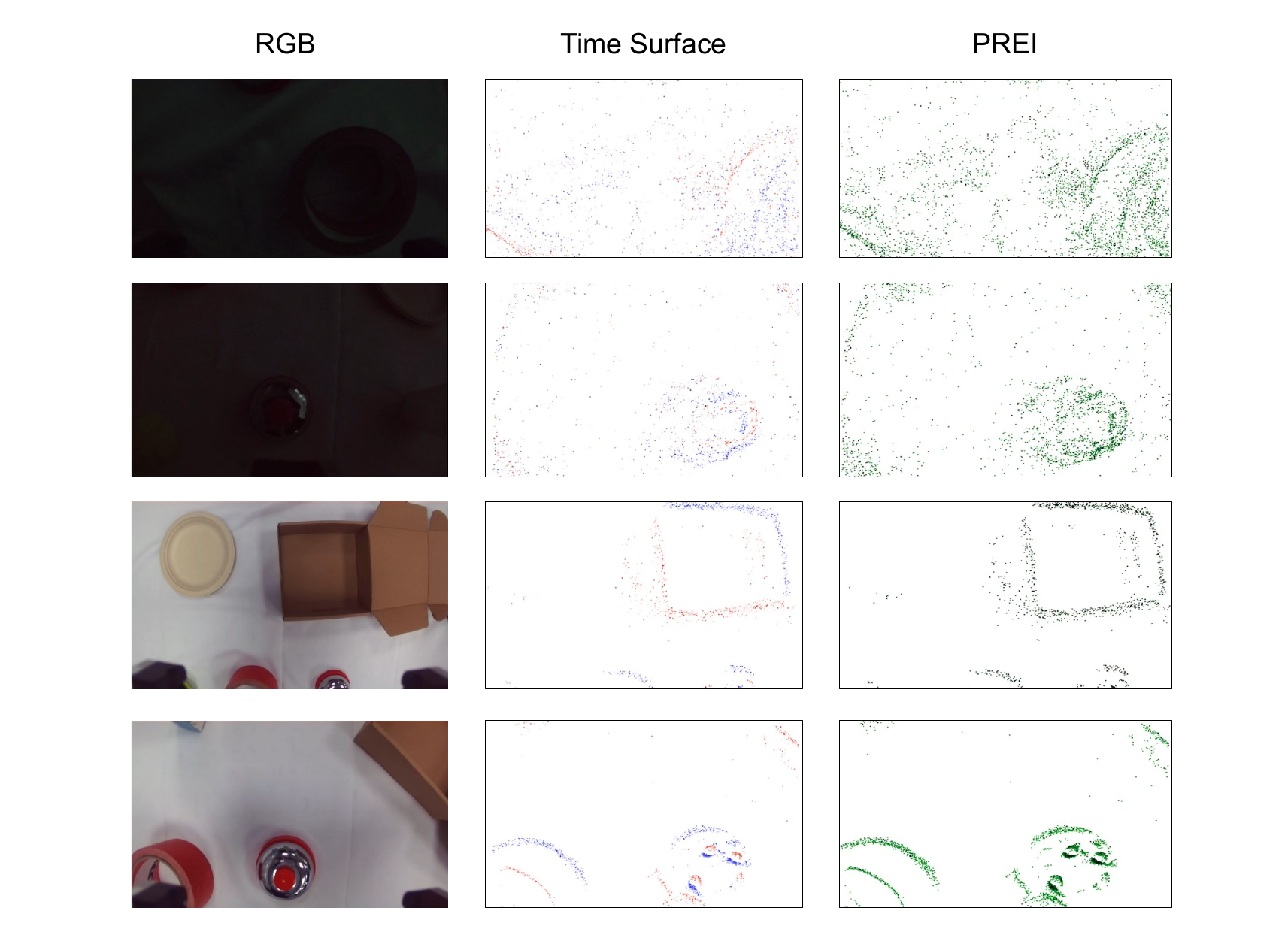}
    \caption{
    Qualitative comparison of time surfaces and PREI under low-light and normal-light conditions.
Compared with time surfaces, PREI provides more stable and structured event representations across illumination changes.
    }
    \label{fig:viz_ts_peri}
\end{figure}

We compare PREI with time surfaces under both low-light and normal-light conditions, as shown in Fig.~\ref{fig:viz_ts_peri}. PREI provides more stable information across different illumination conditions.

\subsection{Event Encoder and Feature Distillation}
\label{app:event_encoder}

Given a PREI map $E_t^{\mathrm{prei}}$, the event encoder extracts patch-level features and projects them into the hidden dimension of the pretrained VLA:
\begin{equation}
U_t=\Phi_e(E_t^{\mathrm{prei}})=\{u_{t,p}\}_{p=1}^{P},
\end{equation}
\begin{equation}
z_{t,p}^e=W_eu_{t,p}\in\mathbb{R}^{d}.
\end{equation}
We use the globally pooled event token and patch-level event tokens as
\begin{equation}
g_t^e=\frac{1}{P}\sum_{p=1}^{P}z_{t,p}^e,\quad
P_t^e=\{z_{t,p}^e\}_{p=1}^{P},
\end{equation}
\begin{equation}
    Y_t^e=[g_t^e;P_t^e].
\end{equation}

To improve feature-level compatibility with the pretrained VLA hidden space, we distill the event encoder from a frozen visual teacher. Given paired RGB observations $I_t$ and PREI maps $E_t^{\mathrm{prei}}$, the teacher produces visual features $Z_t^v=\Phi_T(I_t)=\{z_{t,p}^v\}_{p=1}^{P}$, while the event encoder produces projected event features $Z_t^e=\{z_{t,p}^e\}_{p=1}^{P}$. The distillation objective is
\begin{equation}
\mathcal{L}_{\mathrm{distill}}
=
\lambda_g\left(1-\cos(\bar{z}_t^v,\bar{z}_t^e)\right)
+
\lambda_p\frac{1}{P}\sum_{p=1}^{P}
\left(1-\cos(z_{t,p}^v,z_{t,p}^e)\right),
\end{equation}
where $\Phi_T$ is frozen, and $\bar{z}_t^v$ and $\bar{z}_t^e$ denote globally pooled RGB and event features, respectively. This distillation aligns event features with the VLA-compatible hidden space, while keeping event tokens produced by a separate event encoder outside the global RGB-instruction semantic context of the pretrained VLA.

\begin{figure}
    \centering
    \includegraphics[width=1\linewidth]{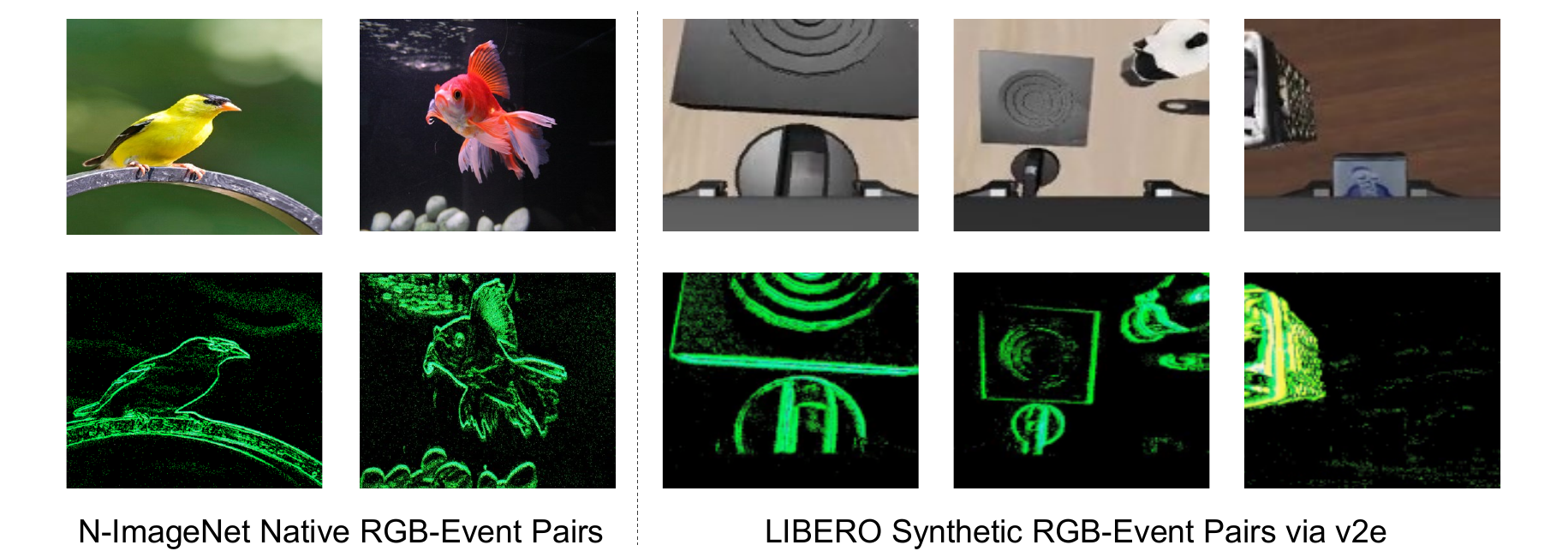}
    \caption{
    Examples of RGB-event pairs used for feature distillation. 
Left: native RGB-event pairs from N-ImageNet. 
Right: synthetic RGB-event pairs generated from LIBERO videos using v2e. 
The top row shows RGB frames, and the bottom row shows the corresponding event representations.
    }
    \label{fig:distillation}
\end{figure}

For feature distillation, we use native RGB-event pairs from N-ImageNet, and synthesize RGB-event pairs from LIBERO videos with v2e, as shown in Fig.~\ref{fig:distillation}. We use the DINO branch as the frozen visual teacher and resize all RGB observations and event representations to $224 \times 224$. The event encoder is trained for 50 epochs with a batch size of 32, learning rate $5\times10^{-5}$, and weight decay $0.01$. The global and patch-level distillation losses are weighted by $\lambda_g=0.7$ and $\lambda_p=0.3$, respectively. 

\textbf{Distillation evaluation metrics.} We further evaluate the quality of feature distillation with four alignment metrics. 
Given $N$ RGB-event pairs, let $\bar{z}_n^v,\bar{z}_n^e\in\mathbb{R}^{d}$ denote the globally pooled RGB and event features, and let $z_{n,p}^v,z_{n,p}^e\in\mathbb{R}^{d}$ denote the $p$-th patch feature of the $n$-th sample. 
The patch-level MSE is computed as
\begin{equation}
\mathrm{MSE}_{\mathrm{patch}}
=
\frac{1}{NP}\sum_{n=1}^{N}\sum_{p=1}^{P}
\left\|z_{n,p}^{v}-z_{n,p}^{e}\right\|_2^2 .
\end{equation}
For cross-modal retrieval, we normalize global RGB and event features and compute the cosine similarity matrix
\begin{equation}
S_{ij}
=
\frac{\langle \bar{z}_i^v,\bar{z}_j^e\rangle}
{\|\bar{z}_i^v\|_2\|\bar{z}_j^e\|_2},
\quad
\mathrm{R@}k
=
\frac{1}{N}\sum_{i=1}^{N}
\mathbb{I}\left[i\in\mathrm{TopK}(S_{i,:},k)\right],
\end{equation}
where RGB$\rightarrow$Event retrieval uses RGB features as queries and event features as keys, and Event$\rightarrow$RGB retrieval is computed analogously. 
Patch Acc@$k$ measures whether the corresponding event patch is retrieved among the top-$k$ nearest patches:
\begin{equation}
\mathrm{PatchAcc@}k
=
\frac{1}{NP}\sum_{n=1}^{N}\sum_{p=1}^{P}
\mathbb{I}\left[
p\in\mathrm{TopK}
\left(
\left\{
\frac{\langle z_{n,p}^{v},z_{n,q}^{e}\rangle}
{\|z_{n,p}^{v}\|_2\|z_{n,q}^{e}\|_2}
\right\}_{q=1}^{P},k
\right)
\right].
\end{equation}
Finally, we compute linear CKA between RGB and event patch features:
\begin{equation}
\mathrm{CKA}(X,Y)
=
\frac{\|X^\top Y\|_F^2}
{\sqrt{\|X^\top X\|_F^2\|Y^\top Y\|_F^2}},
\end{equation}
where $X$ and $Y$ are centered RGB and event patch-feature matrices, respectively.

\begin{table}[htbp]
\centering
\caption{
Feature distillation quality of different event representations on the validation split.
Lower MSE Patch is better, while higher retrieval, patch accuracy, and CKA indicate better RGB-event feature alignment.
}
\label{tab:distill_metrics}
\small
\resizebox{\linewidth}{!}{
\begin{tabular}{lcccc}
\toprule
Event Rep. 
& MSE Patch $\downarrow$ 
& RGB$\rightarrow$Event / Event$\rightarrow$RGB R@1 $\uparrow$ 
& Patch Acc@5 $\uparrow$ 
& Patch CKA $\uparrow$ \\
\midrule
Time Surface & 29.202 & 73.55 / 67.77 & 91.68 & 94.15 \\
PREI & \textbf{26.870} & \textbf{81.36 / 75.91} & \textbf{94.25} & \textbf{95.33} \\
\bottomrule
\end{tabular}
}
\end{table}

We split all RGB-event pairs into training and validation sets with an 8:2 ratio. 
The event encoder is trained on the training split, and all alignment metrics are reported on the held-out validation split. 
As shown in Table~\ref{tab:distill_metrics}, PREI consistently outperforms time surfaces across all metrics, achieving lower patch-level regression error, higher bidirectional RGB-event retrieval accuracy, better local patch correspondence, and stronger representational similarity. 
These results indicate that PREI provides event features that are more compatible with the RGB-based visual teacher and the pretrained VLA hidden space. 

\subsection{Action-Conditioned Routing Details}
\label{app:routing_details}

This section provides the full formulation of the post-backbone routing
interface used in Sec.~3.3. The learnable query tokens are divided into
common queries $Q_0^c$, action queries $Q_0^a$, and event queries $Q_0^e$.
Common queries provide task-level context, action queries condition
action-oriented routing, and event queries support event-oriented routing
and the auxiliary future-PREI prediction objective.

The input to the pretrained VLA backbone is
\begin{equation}
X_t = [Z_t^v; z_t^s; Z^\ell; Z_0^a; Q_0^c; Q_0^a; Q_0^e],
\end{equation}
where $Z_t^v$ denotes RGB visual tokens, $z_t^s$ denotes the proprioceptive
token, $Z^\ell$ denotes language tokens, and $Z_0^a$ denotes action
placeholders. Event tokens are not included in $X_t$.

The VLA backbone produces
\begin{equation}
H_t = F_{\mathrm{vla}}(X_t), \qquad
H_t = [Z_t^{\mathrm{lm}}; Q_t^c; Q_t^a; Q_t^e],
\end{equation}
where $Z_t^{\mathrm{lm}}$ contains the contextualized RGB, proprioceptive,
language, and action-placeholder states.

Given the encoded PREI event tokens $Y_t^e$, event information is injected
after the VLA backbone:
\begin{equation}
Z_t^{\mathrm{attn}}
=
\mathrm{CrossAttn}
\left(
W_q Z_t^{\mathrm{lm}},
W_k (Y_t^e + T_{\mathrm{pos}}),
W_v (Y_t^e + T_{\mathrm{pos}})
\right).
\end{equation}

The fusion gate is computed as
\begin{equation}
\Gamma_t =
\sigma
\left(
f_{\mathrm{gate}}
(
Z_t^{\mathrm{lm}}, Q_t^c, Q_t^a, Q_t^e, Z_t^{\mathrm{attn}}
)
\right),
\end{equation}
and the fused hidden tokens are
\begin{equation}
Z_t^{\mathrm{fused}}
=
\mathrm{Norm}
\left(
Z_t^{\mathrm{lm}} + \Gamma_t \odot Z_t^{\mathrm{attn}}
\right).
\end{equation}

The common and action queries route action-relevant tokens to the action
head, while the common and event queries route event-relevant tokens to the
auxiliary event head:
\begin{equation}
Z_t^{\mathrm{act}}
=
\mathrm{Route}_{\mathrm{act}}
(
Z_t^{\mathrm{fused}}, Q_t^c, Q_t^a
),
\qquad
Z_t^{\mathrm{evt}}
=
\mathrm{Route}_{\mathrm{evt}}
(
Z_t^{\mathrm{fused}}, Q_t^c, Q_t^e
).
\end{equation}
Although the routing operators can be instantiated using different token-selection or aggregation strategies, we use a simple parameter-free implementation that concatenates the fused tokens with the corresponding query tokens. 

The action head predicts the future action chunk, and the event head predicts future PREI event targets conditioned on the routed event tokens and the action representation:
\begin{equation}
    \hat{A}_t
    =
    f_{\mathrm{act}}(Z_t^{\mathrm{act}}),
    \qquad
    \hat{E}^{\mathrm{prei}}_{t+1:t+K}
    =
    f_{\mathrm{evt}}(Z_t^{\mathrm{evt}},\hat{A}_t).
\label{eq:action_event_decode}
\end{equation}

The action head is implemented as a multilayer perceptron, whereas the event head is implemented as a deconvolutional decoder. Since $Y_t^e$ is never passed into $F_{\mathrm{vla}}$, event tokens do not participate in the global semantic self-attention of the pretrained VLA. Instead, they affect action and future-event prediction only through post-backbone gated fusion and query-guided routing.

\subsection{Auxiliary Future Event Objective}
\label{app:auxiliary_event_objective}

The auxiliary future-event objective encourages the fused and event-routed
representations to preserve transient physical information from the event
stream. Given the routed event representation $Z_t^{\mathrm{evt}}$, the
contextualized event queries $Q_t^e$, and the predicted action chunk
$\widehat A_t$, a lightweight event head predicts future PREI targets:
\begin{equation}
    \widehat E_{t+1:t+K}^{\mathrm{prei}}
    =
    f_{\mathrm{evt}}
    (
    Z_t^{\mathrm{evt}}, Q_t^e, \widehat A_t
    ).
\end{equation}

Let $E_{t+k}^{\mathrm{prei}}$ denote the ground-truth future PREI target
and $\widehat E_{t+k}^{\mathrm{prei}}$ denote the corresponding event-head
prediction. We use a content mask
$M_{t+k}\in[0,1]^{H\times W}$ to compute the event loss only on valid
content regions:
\begin{equation}
    \mathcal{L}_{\mathrm{evt}}
    =
    \frac{1}{K}
    \sum_{k=1}^{K}
    \frac{
    \left\|
    M_{t+k}
    \odot
    \left(
    \widehat E_{t+k}^{\mathrm{prei}}
    -
    E_{t+k}^{\mathrm{prei}}
    \right)
    \right\|_1
    }{
    \|M_{t+k}\|_1+\epsilon
    } .
\end{equation}
Here $M_{t+k}$ is broadcast to all PREI channels, and $\epsilon$ is a small
constant for numerical stability.

To further regularize the local event structure, we impose a first-order
derivative consistency loss between the predicted and ground-truth future
PREI maps:
\begin{equation}
    \mathcal{L}_{\mathrm{deriv}}
    =
    \frac{1}{K}
    \sum_{k=1}^{K}
    \frac{
    \left\|
    M_{t+k}
    \odot
    \left(
    \nabla \widehat E_{t+k}^{\mathrm{prei}}
    -
    \nabla E_{t+k}^{\mathrm{prei}}
    \right)
    \right\|_1
    }{
    \|M_{t+k}\|_1+\epsilon
    } ,
\end{equation}
where $\nabla$ denotes the first-order finite-difference derivative operator
on the PREI representation.

The full training objective is
\begin{equation}
    \mathcal{L}
    =
    \mathcal{L}_{\mathrm{act}}
    +
    \lambda_{\mathrm{evt}}\mathcal{L}_{\mathrm{evt}}
    +
    \lambda_{\mathrm{deriv}}\mathcal{L}_{\mathrm{deriv}},
\end{equation}
where we set $\lambda_{\mathrm{evt}}=0.1$ and
$\lambda_{\mathrm{deriv}}=0.3$ in all experiments.

The auxiliary event head is used only as a training-time regularizer.
During inference, action prediction does not require future PREI targets or
event-head supervision.

\section{RGB-to-Event Simulator}
\label{app:rgb2event}

\subsection{RGB-to-Event Simulator}
\label{sec:rgb2event_simulator}

The original LIBERO benchmark does not provide native event streams.
To evaluate event-conditioned policies in a controlled and reproducible
setting, we implement an efficient RGB-to-event simulator that generates
synchronized event-side observations from LIBERO rendered videos.

\paragraph{Pseudo-event supervision.}
We first process LIBERO rendered videos using v2e to obtain pseudo-event
streams. These pseudo-events provide supervision for learning compact
event-side representations. Let
$\mathcal{V}=\{I_t\}_{t=1}^{T}$ denote a LIBERO RGB trajectory and
$\mathcal{S}_{t-H:t}$ denote the corresponding pseudo-event stream over a
causal temporal window. From $\mathcal{S}_{t-H:t}$, we compute compact
event representations, including time surfaces and PREI maps, which serve
as training targets for the simulator. We use PREI as the default event
representation for Event-VLA, while time surfaces are used for
representation ablations.

\paragraph{Compact representation prediction.}
Unlike full event-stream reconstruction, Event-VLA only requires compact
event-side representations. Therefore, we train lightweight pixel-to-pixel
converters to directly predict event representations from RGB observations:
\begin{equation}
    \widehat{E}^{\mathrm{ts}}_t
    =
    f_{\mathrm{ts}}(I_{t-H:t}),
    \qquad
    \widehat{E}^{\mathrm{prei}}_t
    =
    f_{\mathrm{prei}}(I_{t-H:t}),
    \label{eq:rgb2event_prediction}
\end{equation}
where $f_{\mathrm{ts}}$ and $f_{\mathrm{prei}}$ are U-Net-style predictors,
$I_{t-H:t}$ denotes a short causal RGB history,
$\widehat{E}^{\mathrm{ts}}_t$ is the predicted time-surface representation,
and $\widehat{E}^{\mathrm{prei}}_t$ is the predicted PREI representation.

\paragraph{Training objective.}
The simulator is trained with a reconstruction objective over event
representations:
\begin{equation}
    \mathcal{L}_{\mathrm{sim}}
    =
    \lambda_1
    \left\|
    \widehat{E}_t - E_t
    \right\|_1
    +
    \lambda_{\mathrm{grad}}
    \left\|
    \nabla \widehat{E}_t - \nabla E_t
    \right\|_1,
    \label{eq:simulator_loss}
\end{equation}
where $E_t$ denotes the pseudo-event target,
$\widehat{E}_t$ denotes the simulator prediction, and $\nabla$ denotes the
spatial image-gradient operator. The first term encourages pixel-level
fidelity, while the gradient term promotes edge and structural consistency. We set $\lambda_1=1.0$ and $\lambda_{\mathrm{grad}}=0.2$ in all experiments.

\paragraph{Representation-level validation.}
We validate the RGB-to-event simulator by comparing predicted event
representations against pseudo-event targets on held-out LIBERO trajectories.
Let $\widehat{E}$ and $E$ denote the predicted and target event
representations, respectively, and let $N$ be the total number of evaluated
pixels across spatial locations and channels. We report five complementary
metrics: MAE, RMSE, PSNR, MS-SSIM, and Edge-L1.

The mean absolute error is defined as
\begin{equation}
    \mathrm{MAE}
    =
    \frac{1}{N}
    \sum_{i=1}^{N}
    \|
    \widehat{E}_i - E_i
    \|,
    \label{eq:metric_mae}
\end{equation}
which measures the average pixel-wise absolute reconstruction error.

The root mean squared error is defined as
\begin{equation}
    \mathrm{RMSE}
    =
    \sqrt{
    \frac{1}{N}
    \sum_{i=1}^{N}
    \left(
    \widehat{E}_i - E_i
    \right)^2
    },
    \label{eq:metric_rmse}
\end{equation}
which penalizes large reconstruction errors more strongly than MAE.

The peak signal-to-noise ratio is defined as
\begin{equation}
    \mathrm{PSNR}
    =
    10 \log_{10}
    \left(
    \frac{L^2}{\mathrm{MSE}}
    \right),
    \qquad
    \mathrm{MSE}
    =
    \frac{1}{N}
    \sum_{i=1}^{N}
    \left(
    \widehat{E}_i - E_i
    \right)^2,
    \label{eq:metric_psnr}
\end{equation}
where $L$ denotes the maximum possible signal value of the evaluated
representation. When event representations are normalized to $[0,1]$, we use
$L=1$. Higher PSNR indicates better reconstruction fidelity.

The multi-scale structural similarity index is defined as
\begin{equation}
    \mathrm{MS\text{-}SSIM}
    =
    \left[
    l_M(\widehat{E}, E)
    \right]^{\alpha_M}
    \prod_{j=1}^{M}
    \left[
    c_j(\widehat{E}, E)
    \right]^{\beta_j}
    \left[
    s_j(\widehat{E}, E)
    \right]^{\gamma_j},
    \label{eq:metric_msssim}
\end{equation}
where $l_M$, $c_j$, and $s_j$ denote luminance, contrast, and structure
similarities computed across multiple scales. MS-SSIM measures structural
consistency beyond point-wise pixel errors.

\begin{figure}[htb]
    \centering
    \includegraphics[width=1\linewidth]{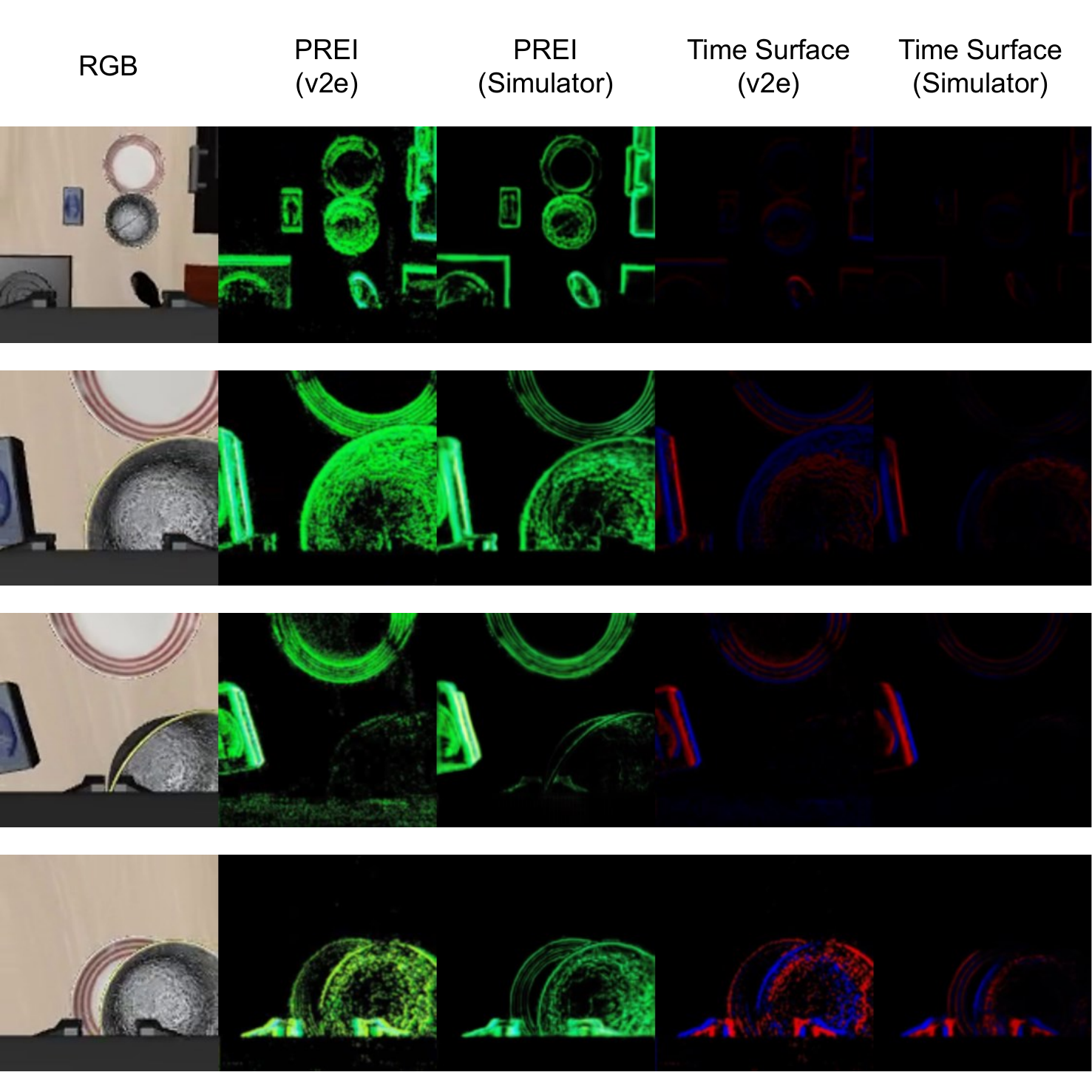}
    \caption{
    Qualitative visualization of RGB-to-event simulator outputs on
held-out LIBERO trajectories. From left to right, we show the input RGB frame,
the v2e-generated PREI target, the simulator-predicted PREI, the v2e-generated
time surface target, and the simulator-predicted time surface. The simulator
recovers the main event structures of the v2e targets while directly producing
compact event-side representations without generating raw event streams.
    }
    \label{fig:simulator_viz}
\end{figure}

Finally, we report an edge consistency metric:
\begin{equation}
    \mathrm{Edge\text{-}L1}
    =
    \frac{1}{N}
    \sum_{i=1}^{N}
    \|
    \nabla \widehat{E}_i - \nabla E_i
    \|,
    \label{eq:metric_edge_l1}
\end{equation}
where $\nabla$ is implemented using a Sobel gradient operator. Edge-L1
directly evaluates whether the simulator preserves the edge and contour
structures that are critical for event-side observations.
\begin{table*}[htb]
\centering
\caption{RGB-to-event simulator validation on held-out LIBERO trajectories.
Lower MAE, RMSE, and Edge-L1 indicate smaller reconstruction errors, while
higher PSNR and MS-SSIM indicate better fidelity and structural consistency.}
\label{tab:sim_validation}
\setlength{\tabcolsep}{8pt}
\renewcommand{\arraystretch}{1.05}
\begin{tabular*}{\textwidth}{@{\extracolsep{\fill}}lccccc}
\toprule
Representation
& MAE $\downarrow$
& RMSE $\downarrow$
& PSNR $\uparrow$
& MS-SSIM $\uparrow$
& Edge-L1 $\downarrow$ \\
\midrule
Time surface & 0.028 & 0.085 & 21.483 & 0.780 & 0.101 \\
PREI
& \textbf{0.008}
& \textbf{0.043}
& \textbf{27.858}
& \textbf{0.897}
& \textbf{0.031} \\
\bottomrule
\end{tabular*}
\end{table*}

\paragraph{Use during evaluation.}
During LIBERO-Cross evaluation, we use the trained RGB-to-event simulator to
directly convert RGB frames into compact event-side representations. This avoids
the expensive v2e pipeline, which requires interpolating RGB videos and then
generating raw event streams. The simulator therefore provides an efficient way
to obtain synchronized event observations for evaluating event-conditioned
policies under visual degradation. To qualitatively assess the simulator outputs, we also visualize the
v2e-generated event representations and the corresponding simulator predictions
in Figure~\ref{fig:simulator_viz}.

\paragraph{Downstream validation.}
In addition to representation-level validation, we further evaluate whether
simulator-produced event representations preserve their utility for downstream
policy execution. As shown in Table~\ref{tab:sim_validation}, PREI achieves
better reconstruction fidelity than time surfaces across all representation-level
metrics, including lower MAE, RMSE, and Edge-L1, as well as higher PSNR and
MS-SSIM. We then compare Event-VLA performance with different simulator-produced
event representations under the challenging LL-Severe degradation setting.
Table~\ref{tab:sim_downstream} shows that PREI also leads to consistently higher
success rates across all LIBERO-Cross task categories. These results indicate
that PREI is not only reconstructed more accurately by the simulator, but also
provides stronger event-side information for robust policy execution.

\paragraph{Simulator leakage control.}
Since LIBERO does not provide native event streams, LIBERO-Cross uses a
RGB-to-event simulator to produce event-side observations. To avoid privileged
information leakage, the simulator is trained independently from the downstream
policy and is frozen during policy training and evaluation. It is conditioned
only on visual observations, and does not use language instructions, robot
states, ground-truth actions, task identities, success labels, or evaluation
outcomes. The same simulator is used for all event-conditioned variants in our
representation and interface comparisons, so these comparisons differ only in
the event representation or the event-to-VLA interface. At each policy step, the simulator uses only past and current visual frames and does not access future frames.
The event-side observation is generated from the same underlying trajectory as
the degraded RGB observation, corresponding to an additional event sensor
observing the same robot-object motion, rather than to privileged task-level
annotations.

\begin{table*}[htbp]
\centering
\caption{Downstream validation of simulator-produced event representations.
Success rate (\%) is reported on LIBERO-Cross under the LL-Severe setting.}
\label{tab:sim_downstream}
\setlength{\tabcolsep}{12pt}
\renewcommand{\arraystretch}{1.05}
\begin{tabular*}{\textwidth}{@{\extracolsep{\fill}}lccccc}
\toprule
Event representation & Spatial & Object & Goal & Long & Avg. \\
\midrule
Time surface & 90.4 & 92.6 & 93.2 & 88.4 & 91.15 \\
PREI
& \textbf{95.6}
& \textbf{97.2}
& \textbf{97.8}
& \textbf{92.0}
& \textbf{95.65} \\
\bottomrule
\end{tabular*}
\end{table*}

\section{Details of LIBERO-Cross}
\label{app:libero_cross}

\begin{table*}[t]
\centering
\caption{Progressive visibility degradation levels in LIBERO-Cross.}
\label{tab:degradation_levels}
\setlength{\tabcolsep}{12pt}
\renewcommand{\arraystretch}{1.05}
\begin{tabular*}{\textwidth}{@{\extracolsep{\fill}}lccc}
\toprule
Level & Exposure reduction / gain @18\% gray & SNR @18\% gray & Motion blur \\
\midrule
LL-Mild   & 2.32 EV / -6.9 dB  & 15.5 dB  & 1 px  \\
LL-Dark   & 3.24 EV / -9.7 dB & -0.2 dB  & 2 px \\
LL-Severe & 4.78 EV / -14.4 dB & -11.9 dB & 3 px \\
\bottomrule
\end{tabular*}
\end{table*}

LIBERO-Cross applies controlled low-light degradations to RGB observations
while keeping the task, language instruction, robot state, and evaluation
protocol unchanged from LIBERO. Each degradation level combines luminance
attenuation, signal-dependent sensor noise, and motion blur. The
degradation levels are summarized in Table~\ref{tab:degradation_levels}.

\subsection{Low-Light Degradation Model}
\label{app:degradation_model}

\paragraph{Luminance attenuation.}
We measure low-light severity at a reference $18\%$ gray level. Given an
exposure factor $e$ and gamma value $\gamma$, and ignoring black-floor and
minimum-luminance clipping terms for the scalar severity summary, the
relative luminance ratio at $18\%$ gray is
\begin{equation}
    r_{18}=e \cdot 0.18^{\gamma-1}.
\end{equation}
The corresponding exposure reduction is reported in exposure value (EV) as
\begin{equation}
    \Delta \mathrm{EV}_{18} = -\log_2(r_{18}).
\end{equation}
We also report the equivalent logarithmic gain in dB:
\begin{equation}
    g_{\mathrm{dB}} = 10\log_{10}(r_{18}).
\end{equation}
Since $r_{18}<1$ under low-light degradation, $g_{\mathrm{dB}}$ is negative.

\paragraph{Signal-dependent noise.}
We use a signal-dependent Gaussian noise model:
\begin{equation}
    \sigma(I)=\sigma_{\mathrm{read}}+\sigma_{\mathrm{shot}}\sqrt{I},
\end{equation}
where $I \in [0,1]$ is the degraded intensity. We summarize the noise level
using the signal-to-noise ratio at $18\%$ gray:
\begin{equation}
    \mathrm{SNR}_{18}
    =
    20\log_{10}
    \frac{I_{18}}{\sigma(I_{18})},
    \quad
    I_{18}=e\cdot 0.18^\gamma.
\end{equation}

\paragraph{Motion blur.}
We apply weak linear motion blur to simulate slight camera or object motion
under low-light imaging. The blur strength is reported as the kernel length
in pixels and as its normalized value with respect to the ViT patch size
$P=16$:
\begin{equation}
    B_{\mathrm{patch}}
    =
    \frac{L_{\mathrm{blur}}}{P}.
\end{equation}
The maximum blur strength is set to $3$ pixels, corresponding to $0.19$ patch.
Since real-world manipulation typically involves moderate motion speeds, motion
blur is less pronounced than the sensor noise introduced by long exposure under
low-light conditions.

\begin{figure}[htbp]
    \centering
    \includegraphics[width=1\linewidth]{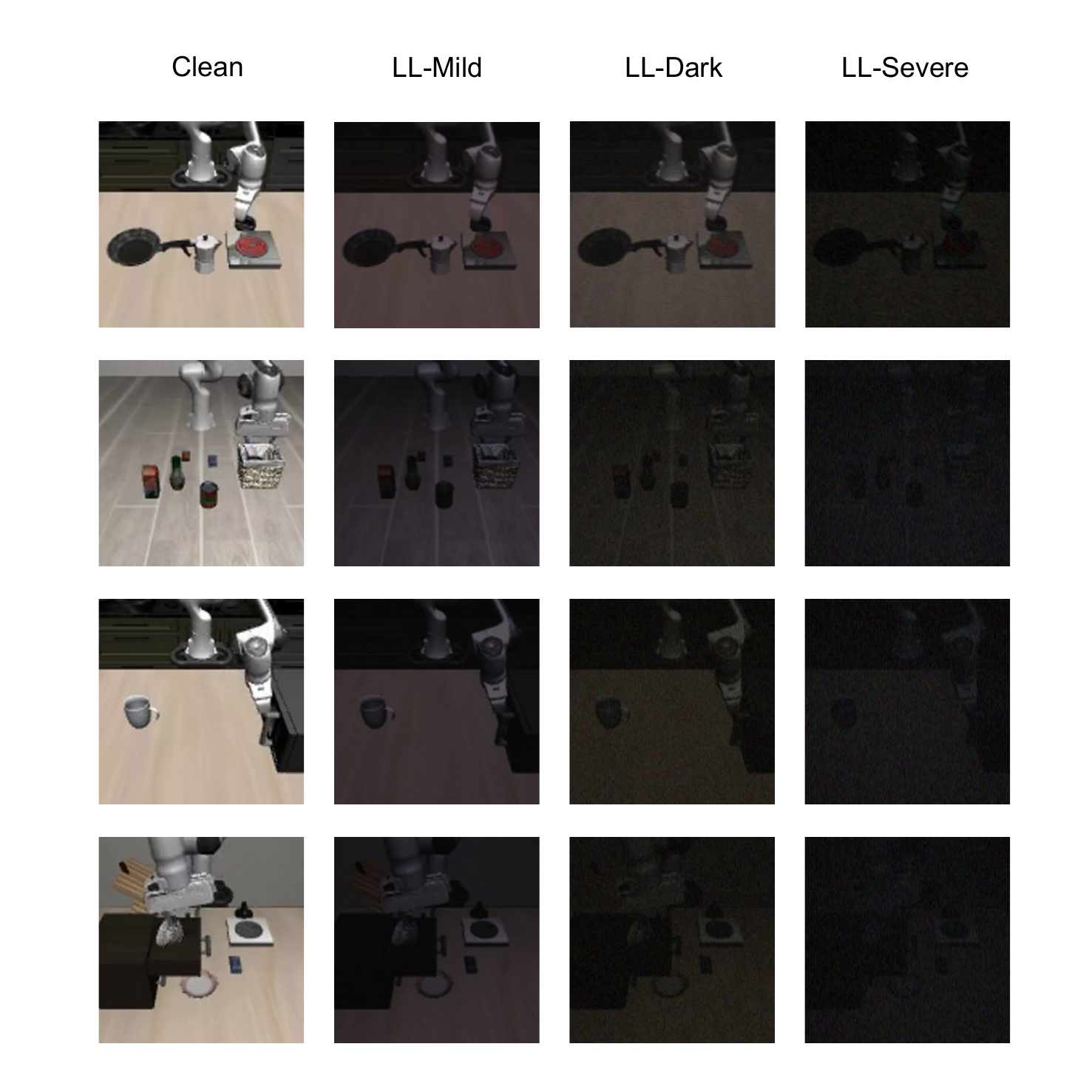}
    \caption{
    Examples of low-light visual degradations in LIBERO-Cross.
Columns show the original RGB observation and progressively stronger
low-light corruptions: LL-Mild, LL-Dark, and LL-Severe.
    }
    \label{fig:libero_cross_examples}
\end{figure}

\subsection{Visualization of Degradation Levels}
\label{app:degradation_visualization}

Figure~\ref{fig:libero_cross_examples} shows example RGB observations from
LIBERO-Cross under the clean setting and three degradation levels: LL-Mild,
LL-Dark, and LL-Severe. As the degradation becomes stronger, frame-based RGB
observations lose texture, boundary, and object-level visual cues, motivating the
use of event-derived residual observations for robust manipulation.

\section{Implementation Details}
\label{app:implementation}

\paragraph{Training details.}
We train Event-VLA separately on each sub-benchmark using the same
training recipe. The policy is initialized from an OpenVLA checkpoint with all other modules are randomly initialized. We use an effective batch size of 64 on
eight H100 GPUs, action chunk size 8, and event history length 8. Each run is
trained for roughly 140k optimization steps.

\paragraph{Optimization.}
We optimize with AdamW using learning rate $5\times10^{-5}$. The learning rate
is decayed by $0.1$ after 50k steps. We use the default AdamW weight decay
($0.01$). Auxiliary event
prediction losses are warmed up for 5k steps during the early stage of training. The pretrained VLA backbone is LoRA-tuned with rank 32.

% \paragraph{Baselines.}
% For RGB-only baselines, we keep the same OpenVLA-OFT initialization, training
% budget, action head, proprioception, LoRA setting, and evaluation protocol,
% but disable event conditioning. For event-interface baselines, we match the
% training budget and pretrained VLA backbone whenever possible. The compared
% event interfaces include global event tokens, RGB/event adapter fusion,
% generic cross-attention fusion, and MM-ACT-style shared-context event fusion.

\paragraph{Evaluation protocol.}
We evaluate each policy in the simulator with the official task success
signal. A rollout is successful if the environment reports task completion
within the task horizon. We use 50 rollouts per
task and report the average success rate over all tasks in the evaluated sub-benchmark. For low-light evaluation, we apply the corresponding
controlled degradation level to RGB observations while keeping the task,
language instruction, robot state, and rollout protocol unchanged.

\section{Additional Interface Ablations}
\label{app:additional_ablations}
\label{app:ablation_degradation}
\label{app:ablation_normal}
\label{app:efficiency}

We provide detailed sub-benchmark results for the ablation study in
Table~\ref{tab:ablation_detailed_subbench}. All ablations are conducted on
LIBERO-Cross under the LL-Severe setting, where RGB observations are most
severely degraded. We report success rate (\%) on each LIBERO-Cross task
category and the average success rate across categories.

\begin{table*}[t]
\centering
\caption{Detailed ablation results on LIBERO-Cross under the LL-Severe setting.
We report success rate (\%) on each task category and the average success rate.}
\label{tab:ablation_detailed_subbench}
\setlength{\tabcolsep}{6pt}
\renewcommand{\arraystretch}{1.05}
\begin{tabular*}{\textwidth}{@{\extracolsep{\fill}}llccccc}
\toprule
Type & Variant & Spatial & Object & Goal & Long & Avg. \\
\midrule
\multirow{3}{*}{Representation}
& No event     & 51.6 & 74.8 & 55.8 & 60.4 & 60.6 \\
& Time surface & 90.4 & 92.6 & 93.2 & 88.4 & 91.2 \\
& PREI         & \textbf{95.6} & \textbf{97.2} & \textbf{97.8} & \textbf{92.0} & \textbf{95.6} \\
\midrule
\multirow{3}{*}{Interface}
% 94.6	97.4	96.8	91.6	95.1
& Unified encoding  & 94.6 & \textbf{97.4} & 96.8 & 91.6 & 95.1 \\
& RGB/event adapter & 93.4 & 96.6 & 96.2 & 90.8 & 94.2 \\
& Query routing     & \textbf{95.6} & 97.2 & \textbf{97.8} & \textbf{92.0} & \textbf{95.6} \\
\midrule
\multirow{3}{*}{Query}
& w/o common query & 94.6 & 96.4 & 95.8 & 91.2 & 94.5 \\
& w/o event query  & \textbf{95.6} & 96.6 & 96.2 & \textbf{92.4} & 95.2 \\
& Full queries     & \textbf{95.6} & \textbf{97.2} & \textbf{97.8} & 92.0 & \textbf{95.6} \\
\midrule
\multirow{3}{*}{Regularization}
& None           & 94.2 & 95.8 & \textbf{97.6} & 91.6 & 94.8 \\
& w/o mask       & 94.8 & 96.2 & 97.2 & \textbf{92.2} & 95.1 \\
& Full objective & \textbf{95.6} & \textbf{97.2} & \textbf{97.8} & 92.0 & \textbf{95.6} \\
\bottomrule
\end{tabular*}
\end{table*}

\paragraph{Representation ablation.}
We first study the effect of event representation. The no-event variant disables
the event pathway and uses only RGB, language, and proprioceptive inputs. Time
surface provides a standard event representation based on recent event activity.
PREI is our default representation, which integrates instantaneous, salient, and
persistent event responses into a compact residual map. Compared with the
no-event baseline, both event representations substantially improve robustness
under LL-Severe. PREI achieves the best average performance, indicating that its
denser and multi-timescale residual representation provides more useful physical
cues for action prediction.

\paragraph{Interface ablation.}
We then compare different ways of incorporating event information into the VLA.
Unified encoding directly concatenates event tokens with RGB, language, and
action tokens inside the pretrained VLA token sequence. Although this achieves
competitive success rate, it introduces large latency overhead because event
tokens participate in the backbone self-attention. The RGB/event adapter uses a
lightweight adapter to fuse RGB and event features, resulting in low latency but
lower success rate. Our query-routing interface keeps event tokens outside the
pretrained semantic backbone and injects them through gated cross-attention and
query-guided routing. This achieves the best average success rate with only
small additional latency.

\paragraph{Query ablation.}
We further ablate the query design. Removing the common query weakens the shared
task-level context used for routing, while removing the event query reduces the
event-specific pathway. The full query design, which uses common, action, and
event queries jointly, achieves the best average success rate. This suggests
that both shared task context and event-specific routing are useful for
extracting action-relevant event information.

\paragraph{Regularization ablation.}
Finally, we study the effect of the auxiliary future-event objective. The
``None'' variant removes event prediction regularization and trains only with
the action prediction loss. The ``w/o mask'' variant uses future-event
supervision without the content mask. The full objective applies masked
future-event reconstruction together with derivative consistency. These
auxiliary losses are used only during training and are not required during
inference. The improvement from the full objective shows that future-event
regularization helps the routed features preserve task-relevant physical
residual information.

\paragraph{Summary.}
Overall, the ablation results show that Event-VLA's robustness comes from the
combination of PREI representation, action-conditioned query routing, and
auxiliary event regularization. PREI provides the strongest event-side
representation, while query routing offers the best trade-off between success
rate and inference latency.

\section{Real-World Setup and Additional Results}
\label{app:real_robot}

\subsection{Hardware and Calibration}
\label{app:real_hardware}

The real-world setup uses a Franka Research 3 robot. Visual observations are collected from a wrist-mounted ZED camera and an external Orbbec camera. Event observations are collected from a DAVIS event camera. All sensors are calibrated and temporally synchronized before data collection.

\subsection{Tasks and Data Collection}
\label{app:real_tasks}

We evaluate on four manipulation tasks:
\texttt{put the red tape into the cardboard box},
\texttt{put the tennis ball into the cardboard box},
\texttt{put the milk into the plate},
and
\texttt{ring the bell}.
Each task contains 20 demonstrations under different lighting conditions. We evaluate each task in \texttt{Normal}, \texttt{Low Light} and \texttt{Near-Dark} condition for 10 trials, totally 40 trials each condition. The model is deployed in a RTX 4090 server.

\subsection{Lighting Conditions Evaluation}
\label{app:real_lighting}

We evaluate policies under three lighting settings: normal light, low light, and near-dark. The near-dark setting is designed to make RGB observations unreliable while preserving motion-induced event responses.

\begin{table*}[htbp]
\centering
\caption{Per-task real-world success rates under different lighting conditions.
Each task is evaluated over 10 trials. We report success rate (\%) for each task
and the average success rate across four tasks.}
\label{tab:real_world_per_task}
\setlength{\tabcolsep}{8pt}
\renewcommand{\arraystretch}{1.05}
\begin{tabular*}{\textwidth}{@{\extracolsep{\fill}}llccccc}
\toprule
Lighting condition & Method & Task 1 & Task 2 & Task 3 & Task 4 & Avg. \\
\midrule
\multirow{3}{*}{Normal}
& $\pi_0$        & \textbf{80.0} & \textbf{70.0} & 70.0 & \textbf{80.0} & \textbf{75.0} \\
& OpenVLA-OFT     & 70.0 & 60.0 & 70.0 & \textbf{80.0} & 70.0 \\
& Ours w/o queries & 60.0 & 60.0 & 80.0 & 80.0 & 70.0 \\
& Ours            & 70.0 & 60.0 & \textbf{80.0} & \textbf{80.0} & 72.5 \\
\midrule
\multirow{3}{*}{Low-light}
& $\pi_0$       & 50.0 & 40.0 & 60.0 & 70.0 & 55.0 \\
& OpenVLA-OFT   & 60.0 & 40.0 & \textbf{70.0} & 60.0 & 57.5 \\
& Ours w/o queries & 70.0 & 50.0 & 70.0 & 70.0 & 65.0 \\
& Ours          & \textbf{70.0} & \textbf{60.0} & \textbf{70.0} & \textbf{80.0} & \textbf{70.0} \\
\midrule
\multirow{3}{*}{Near-dark}
& $\pi_0$       & 30.0 & 0.0  & 20.0 & 10.0 & 15.0 \\
& OpenVLA-OFT   & 20.0 & 10.0 & 10.0 & 10.0 & 12.5 \\
& Ours w/o queries & 50.0 & 30.0 & 50.0 & 50.0 & 45.0 \\
& Ours          & \textbf{60.0} & \textbf{40.0} & \textbf{50.0} & \textbf{60.0} & \textbf{52.5} \\
\bottomrule
\end{tabular*}
\end{table*}

\begin{figure}[htbp]
    \centering
    \includegraphics[width=1\linewidth]{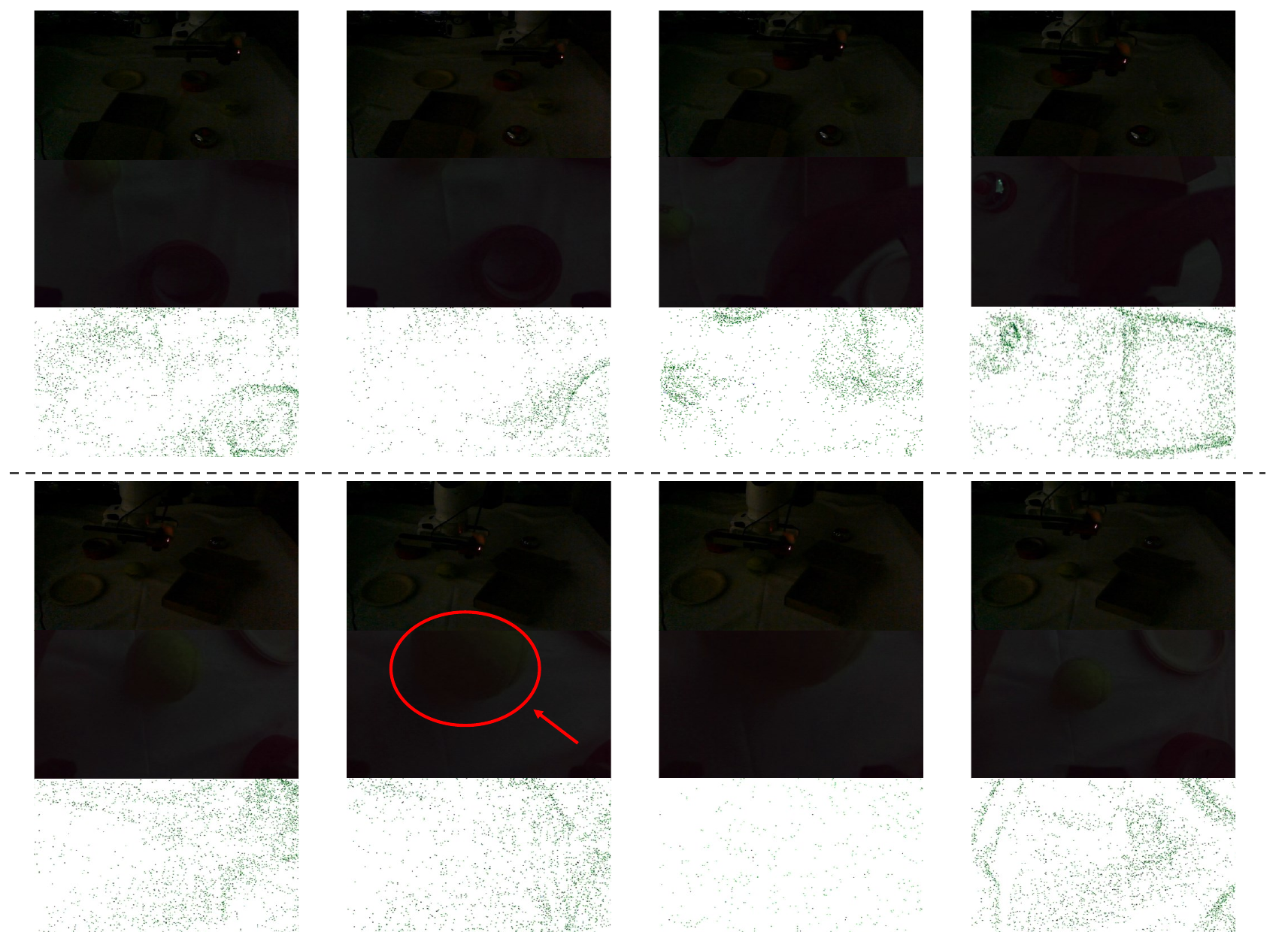}
    \caption{Qualitative real-world sequences under the LL-Severe condition. The
dashed line separates a successful Event-VLA rollout in the upper row from a
failure case in the lower row. In the successful case, event-side observations
provide useful residual cues for manipulation despite severely degraded RGB
frames. In the failure case, the highlighted target region triggers insufficient
event activations, resulting in ambiguous localization and failed manipulation.}
    \label{fig:failure_case}
\end{figure}

\subsection{Qualitative Sequences and Failure Cases}
\label{app:real_qualitative}
Figure~\ref{fig:failure_case} shows representative real-world execution
sequences under the challenging LL-Severe condition. The upper part presents a
successful Event-VLA rollout, where the policy can still exploit event-side
observations to recover task-relevant object and motion cues despite severely
degraded RGB frames. The lower part shows a failure case. In this case, the
target object produces insufficient event activations due to weak contrast,
limited relative motion, and poor illumination. As a result, the event-side
observation provides ambiguous localization cues, causing the policy to
mislocalize the object and fail to complete the manipulation.

These qualitative results show that event-side residual observations improve
robustness under severe low-light degradation, but their effectiveness still
depends on whether the target object generates sufficiently discriminative event
responses.